\documentclass{article}

\usepackage{microtype}
\usepackage{graphicx}
\usepackage{subcaption}
\usepackage{booktabs} 
\usepackage{amsmath}

\usepackage{hyperref}

\usepackage{pdfpages}

\usepackage[preprint]{icml2026}

\usepackage{amsmath}
\usepackage{amssymb}
\usepackage{mathtools}
\usepackage{amsthm}
\usepackage{enumitem}

\usepackage[capitalize,noabbrev]{cleveref}

\usepackage{booktabs}
\usepackage{multirow}
\usepackage{pifont}
\usepackage{xcolor}
\usepackage{makecell}

\usepackage{tcolorbox}
\tcbuselibrary{listings,skins,breakable}
\usepackage{listings}
\lstdefinelanguage{yaml}{
  basicstyle=\ttfamily\footnotesize,
  sensitive=false,
  comment=[l]{\#},
  commentstyle=\color{gray},
  stringstyle=\color{black},
  moredelim=**[is][\bfseries]{**}{**}
}

\newtcblisting{PromptBox}[2][]{%
  breakable,
  enhanced,
  colback=green!4,
  colframe=green!50!black,
  colbacktitle=green!50!black,
  coltitle=white,
  fonttitle=\bfseries,
  title={#2},
  boxed title style={
    sharp corners,
    boxrule=0pt,
    left=2mm, right=2mm, top=1mm, bottom=1mm
  },
  boxrule=0.7pt,
  arc=2mm,
  left=4mm, right=4mm, top=2mm, bottom=2mm,
  listing only,
  listing options={
    language=yaml,
    keepspaces=true
  },
  #1
}
\newtcblisting{PythonBox}[2][]{%
  breakable,
  enhanced,
  colback=blue!3,
  colframe=blue!55!black,
  colbacktitle=blue!55!black,
  coltitle=white,
  fonttitle=\bfseries,
  title={#2},
  boxed title style={
    sharp corners,
    boxrule=0pt,
    left=2mm, right=2mm, top=1mm, bottom=1mm
  },
  boxrule=0.7pt,
  arc=2mm,
  left=4mm, right=4mm, top=2mm, bottom=2mm,
  listing only,
  listing options={
    language=Python
  },
  #1
}

\definecolor{mygreen}{RGB}{0,128,128} 
\definecolor{myred}{RGB}{200,0,0}

\newcommand{\cmark}{\raisebox{0.15ex}{\textcolor{mygreen}{\ding{51}}}} 
\newcommand{\xmark}{\raisebox{0.15ex}{\textcolor{myred}{\ding{55}}}}   

\theoremstyle{plain}

\theoremstyle{definition}

\theoremstyle{remark}

\usepackage[textsize=tiny]{todonotes}

\icmltitlerunning{Submission and Formatting Instructions for ICML 2026}

\begin{document}

\twocolumn[
  \icmltitle{AutoHealth: An Uncertainty-Aware Multi-Agent System \\for Autonomous Health Data Modeling}

    \icmlsetsymbol{equal}{*}
    \icmlsetsymbol{cor}{$\dagger$}

    \begin{icmlauthorlist}
    \icmlauthor{Tong Xia}{equal,cor,1}
    \icmlauthor{Weibin Li}{equal,2}
    \icmlauthor{Gang Liu}{2}
    \icmlauthor{Yong Li}{cor,2}
    \end{icmlauthorlist}
    
    \icmlaffiliation{1}{Vanke School of Public Health, Tsinghua University}
    \icmlaffiliation{2}{Department of Electronic Engineering, Beijing National Research Center for Information Science and Technology, Tsinghua University, China}

    \icmlcorrespondingauthor{Tong Xia}{tongxia@tsinghua.edu.cn}
    \icmlcorrespondingauthor{Yong Li}{liyong07@tsinghua.edu.cn}

    
      \vskip 0.3in
]

\printAffiliationsAndNotice{\icmlEqualContribution}

\begin{abstract}
LLM-based agents have demonstrated strong potential for autonomous machine learning, yet their applicability to health data remains limited. Existing systems often struggle to generalize across heterogeneous health data modalities, rely heavily on predefined solution templates with insufficient adaptation to task-specific objectives, and largely overlook uncertainty estimation, which is essential for reliable decision-making in healthcare. To address these challenges, we propose \textit{AutoHealth}, a novel uncertainty-aware multi-agent system that autonomously models health data and assesses model reliability. \textit{AutoHealth} employs closed-loop coordination among five specialized agents to perform data exploration, task-conditioned model construction, training, and optimization, while jointly prioritizing predictive performance and uncertainty quantification. Beyond producing ready-to-use models, the system generates comprehensive reports to support trustworthy interpretation and risk-aware decision-making. To rigorously evaluate its effectiveness, we curate a challenging real-world benchmark comprising 17 tasks across diverse data modalities and learning settings. \textit{AutoHealth} completes all tasks and outperforms state-of-the-art baselines by 29.2\% in prediction performance and 50.2\% in uncertainty estimation.
\end{abstract}

\vspace{-5mm}
\section{Introduction}



\begin{figure}[t]
    \centering
    \includegraphics[width=0.49\textwidth]{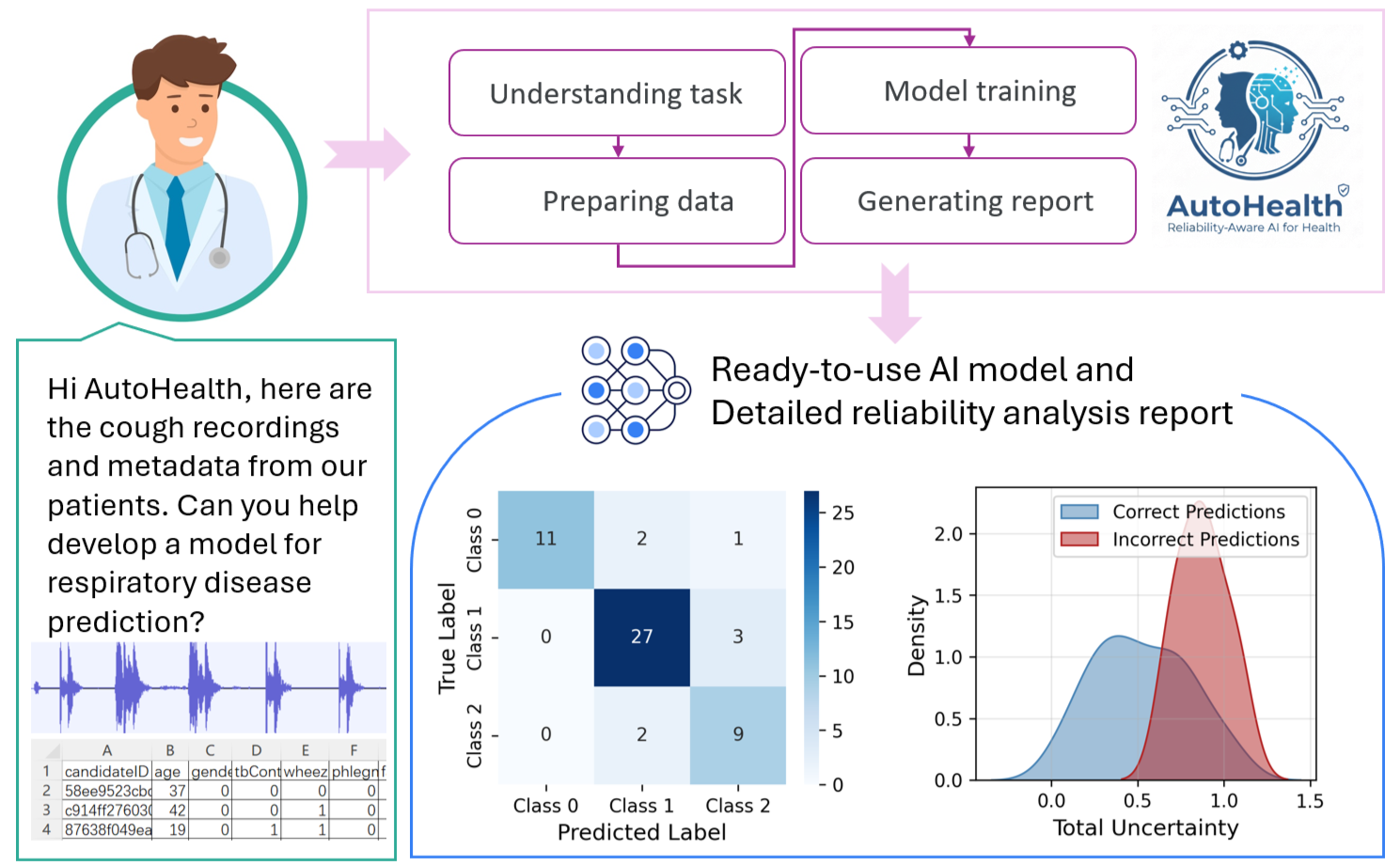}
    \caption{\textit{AutoHealth overview.} A multi-agent system that autonomously constructs predictive models and reliability-aware reports from raw health data.}
    \label{fig:1}
    \vspace{-8mm}
\end{figure}

AI-powered health prediction plays a critical role in disease risk assessment, personalized healthcare, and public health management~\cite{purba2019prediction, rajpurkar2022ai}. However, building reliable prediction models typically requires close collaboration between health experts (e.g., clinicians) and AI experts (e.g., data scientists) to analyze heterogeneous data, design task-specific models, and interpret results. In practice, such collaboration is often hindered by data-sharing constraints and the difficulty of aligning cross-disciplinary expertise, substantially limiting research efficiency and scalability.

Recent advances in large language models (LLMs) and agent-based AI systems, particularly automated data science agents~\cite{chen2025large, Hong2025D, jing2024dsbench}, have begun to reshape scientific workflows by integrating reasoning, coding, and experimentation. These systems promise to lower technical barriers and reduce human effort in model development, making them a promising direction for autonomous health data modeling.

Despite this potential, existing agentic systems remain insufficient when applied to health data. General-purpose systems such as \textit{Claude Code} require careful prompt engineering and substantial technical expertise, making them difficult for health experts to use directly, while end-to-end data science prototypes such as \textit{DS-Agent}~\cite{guo2024ds} exhibit fundamental limitations in healthcare settings (see Sec.~\ref{sec2}). 
First, these systems provide limited support for multi-source, multi-modal, and non–machine-learning–friendly health data, making it difficult to capture the heterogeneity and irregularity inherent in real-world healthcare environments. 
Second, they tend to rely on predefined pipelines or templated solutions, lacking task-conditioned reasoning mechanisms that can flexibly adapt modeling strategies to task-specific objectives and data characteristics. 
Third, existing agents primarily optimize predictive accuracy, while reliability-related aspects, such as interpretability and uncertainty quantification, are rarely treated as first-class objectives, limiting their applicability to high-stakes clinical and public health decision-making.

To address these challenges, we propose \textit{AutoHealth}, an uncertainty-aware multi-agent system for autonomous health data modeling. As illustrated in Figure~\ref{fig:1}, \textit{AutoHealth} accepts high-level modeling objectives and raw health data as input, and autonomously performs task understanding, data preparation, model construction, and evaluation. It produces both deployable predictive models and structured experimental reports, enabling reliable and reproducible health data modeling with minimal human intervention.

Methodologically, \textit{AutoHealth} introduces a closed-loop multi-agent learning paradigm that explicitly targets the key limitations of existing agentic systems. To handle complex and heterogeneous health data, a dedicated data exploration agent performs code-driven analysis and transformation, enabling robust downstream learning across diverse modalities. To balance knowledge reuse with task-specific adaptation, a central coordination agent orchestrates model design and implementation in an adaptive feedback loop, allowing modeling strategies to be iteratively refined based on task objectives and data-derived insights rather than rigid template reuse. Importantly, to ensure model reliability beyond predictive accuracy, \textit{AutoHealth} treats uncertainty quantification as a co-first-class optimization and evaluation objective, with all experimental decisions and results systematically logged and summarized into structured, application-oriented analysis reports to support trustworthy and risk-aware use.

To rigorously evaluate autonomous health data modeling under heterogeneous data conditions and reliability constraints, we construct a challenging real-world benchmark comprising 17 health prediction tasks spanning six data modalities (tabular, image, time series, free text, audio, and graph) and six learning settings (classification, regression, segmentation, survival analysis, forecasting, and link prediction).  Extensive experiments on this benchmark demonstrate that \textit{AutoHealth} consistently outperforms strong baselines, achieving a 100\% task success rate, a 29.2\% improvement in predictive performance, and a 50.2\% improvement in uncertainty quantification.

Our contributions are summarized as follows:
\begin{itemize}
    \item We propose \textit{AutoHealth}, a closed-loop, uncertainty-aware agentic system that autonomously constructs predictive models and reliability-aware analysis reports from raw health data.
    \item We introduce a multi-agent coordination paradigm that enables task-conditioned reasoning and adaptive model construction beyond template-based automated pipelines for heterogeneous health data.
    \item We construct a challenging real-world health prediction benchmark and conduct extensive evaluations, demonstrating that \textit{AutoHealth} achieves consistent and substantial improvements over strong baselines. Benchmark and code are available from \url{https://anonymous.4open.science/r/AutoHealth-46E0}.
\end{itemize}

\begin{table*}[t]
\centering
\caption{Comparison of \textit{AutoHealth} with representative automated and LLM-based data science frameworks.}
\label{tab:comparison}
\scriptsize 
\setlength{\tabcolsep}{4pt} 
\begin{tabular}{lccccccc}
\toprule
\multirow{2}{*}{\textbf{Framework}} & \multicolumn{6}{c}{\textbf{Key Functionality}} \\
\cmidrule(lr){2-8} 
 &\textbf{Multi-agent} & \textbf{Various modalities} & \textbf{Data exploration} & \textbf{With retrieval} & \textbf{Task feedback} & \textbf{Uncertainty} & \textbf{Report} \\
\midrule \midrule 
AutoML-GPT~\cite{Zhang2023AutoMLGPT} & \xmark    & \cmark & \xmark & \xmark & \cmark & \xmark & \xmark \\
HuggingGPT~\cite{Shen2023}           & \xmark    & \cmark & \xmark & \cmark & \xmark & \xmark & \cmark \\
Prompt2Model~\cite{viswanathan2023}  & \xmark    & \xmark & \xmark & \cmark & \xmark & \xmark & \xmark \\

DS-Agent~\cite{guo2024ds}            & \cmark    & \cmark & \xmark & \cmark & \xmark & \xmark & \xmark \\
Data-Interpreter~\cite{Hong2025D}    & \cmark    & \cmark & \cmark & \xmark & \cmark & \xmark & \cmark \\
AutoML-Agent~\cite{Trirat2025}       & \cmark    & \cmark & \xmark & \cmark & \cmark & \xmark & \xmark \\

AutoMMLab~\cite{Yang2025AutoMMLab}   & \cmark    & \xmark & \cmark & \xmark & \xmark & \xmark & \cmark \\
AutoKaggle~\cite{li2024autokaggle}   & \cmark    & \xmark & \cmark & \cmark & \cmark & \xmark & \cmark \\

OpenLens AI~\cite{ChengSuo2025}     & \cmark    & \cmark & \cmark & \cmark & \xmark & \xmark & \cmark \\

\midrule
\textbf{\textit{AutoHealth} (Ours)}  & \cmark & \cmark & \cmark & \cmark & \cmark & \cmark & \cmark\\
\bottomrule
 \vspace{-7mm}
\end{tabular}
\end{table*}

\section{Related Work}\label{sec2}

\subsection{AI for Health}

AI techniques have been widely applied to health-related prediction tasks, including disease risk assessment, diagnostic support, treatment outcome forecasting, and pandemic outbreak prediction~\cite{rajpurkar2022ai, miotto2016deep, Singh2023Pandemic}. These approaches typically leverage heterogeneous data sources such as electronic health records, medical imaging, clinical notes, and longitudinal time-series data to model health outcomes. Despite their success, significant challenges remain, as health data are often heterogeneous, high-dimensional, and incomplete, with substantial variability across institutions and populations~\cite{shickel2018deep}. More fundamentally, most existing approaches remain \emph{task-specific} and \emph{data-dependent}, relying on expert-designed architectures, handcrafted features, modality-specific preprocessing, and domain-informed assumptions. This heavy dependence on manual customization limits transferability and generalization across datasets, modalities, and clinical tasks, motivating the need for more adaptive and automated modeling frameworks to support scalable health data analysis in real-world settings.

\subsection{Reliability-aware Machine Learning}

In high-stakes domains such as healthcare, it is critical not only for models to achieve strong predictive performance but also to provide calibrated uncertainty estimates when deployed in decision support settings~\cite{kompa2021second, ghandeharioun2022distributionaware, xia2024uncertainty}. Uncertainty information can help identify ambiguous or out-of-distribution inputs, enable selective prediction, and support safer human–AI collaboration. Prior work has explored a range of uncertainty quantification techniques for AI, including Bayesian neural networks~\cite{gal2016dropout}, deep ensembles~\cite{lakshminarayanan2017simple}, and test-time data augmentation~\cite{wang2019aleatoric}. However, many health prediction models still lack built-in uncertainty estimation mechanisms, and evaluation protocols often focus primarily on accuracy while overlooking calibration, coverage, or abstention behavior~\cite{carvalho2023machine}.  Overall, prior work underscores the importance of uncertainty quantification for reliable AI-based health prediction, although it is still frequently under-emphasized in model development and evaluation.

\subsection{LLM-based Data Science Agents}

Recent LLMs, such as \textit{Claude Sonnet 4} and \textit{ChatGPT-5}, have demonstrated strong capabilities in code generation and program assistance, enabling the emergence of specialized coding agents such as \textit{Claude Code}. While these tools can substantially accelerate software development, their application to health data modeling often requires careful prompt engineering, environment configuration, and methodological expertise, limiting their accessibility to non-technical health researchers.

In parallel, \textit{automated data science} has emerged as an active research direction, with early efforts focusing on single LLM-based agents that translate natural language instructions into executable modeling pipelines. Representative systems such as \textit{AutoML-GPT}~\cite{Zhang2023AutoMLGPT}, \textit{HuggingGPT}~\cite{Shen2023}, and \textit{Prompt2Model}~\cite{viswanathan2023} established end-to-end workflows integrating data preprocessing, model selection, and evaluation. Despite their conceptual significance, these approaches are inherently constrained by the limited context and reasoning capacity of a single LLM, making it difficult to handle complex, real-world health data and modeling objectives.

To address these limitations, recent work has shifted toward \textit{multi-agent LLM systems}, in which agents with specialized roles collaborate to decompose and solve complex tasks. \textit{DS-Agent}~\cite{guo2024ds} introduces case-based retrieval to reuse successful modeling experiences, improving robustness and task completion, while \textit{Data-Interpreter}~\cite{Hong2025D} emphasizes trajectory-level feedback for iterative refinement based on intermediate execution results. \textit{AutoML-Agent}~\cite{Trirat2025} integrates retrieval and feedback within a unified framework, but primarily targets benchmark-style datasets and conventional optimization objectives, offering limited support for heterogeneous health data and rarely incorporating explicit uncertainty quantification.

Parallel to \textit{general-purpose} automation frameworks, several systems target \textit{domain-specific} scenarios. \textit{AutoMMLab}~\cite{Yang2025AutoMMLab} focuses on computer vision pipelines, while \textit{AutoKaggle}~\cite{li2024autokaggle} is tailored to competition-style data science workflows. Although effective within their respective domains, their task assumptions and pipeline designs limit generalization to broader machine learning and healthcare research settings. \textit{OpenLens AI}~\cite{ChengSuo2025} is most closely aligned with health-related applications, emphasizing exploratory health informatics analysis and visual reporting. However, it provides limited support for systematic machine learning experiment design and does not explicitly incorporate reliability-aware modeling or uncertainty quantification.

We summarize representative prior work in Table~\ref{tab:comparison}. Overall, existing systems typically address isolated aspects of automated modeling, such as pipeline reuse, or feedback refinement, but rarely integrate heterogeneous data exploration, task-conditioned model construction, and uncertainty-aware evaluation within a unified agentic framework.

   \begin{figure*}[t]
    \centering
    \includegraphics[width=0.99\textwidth]{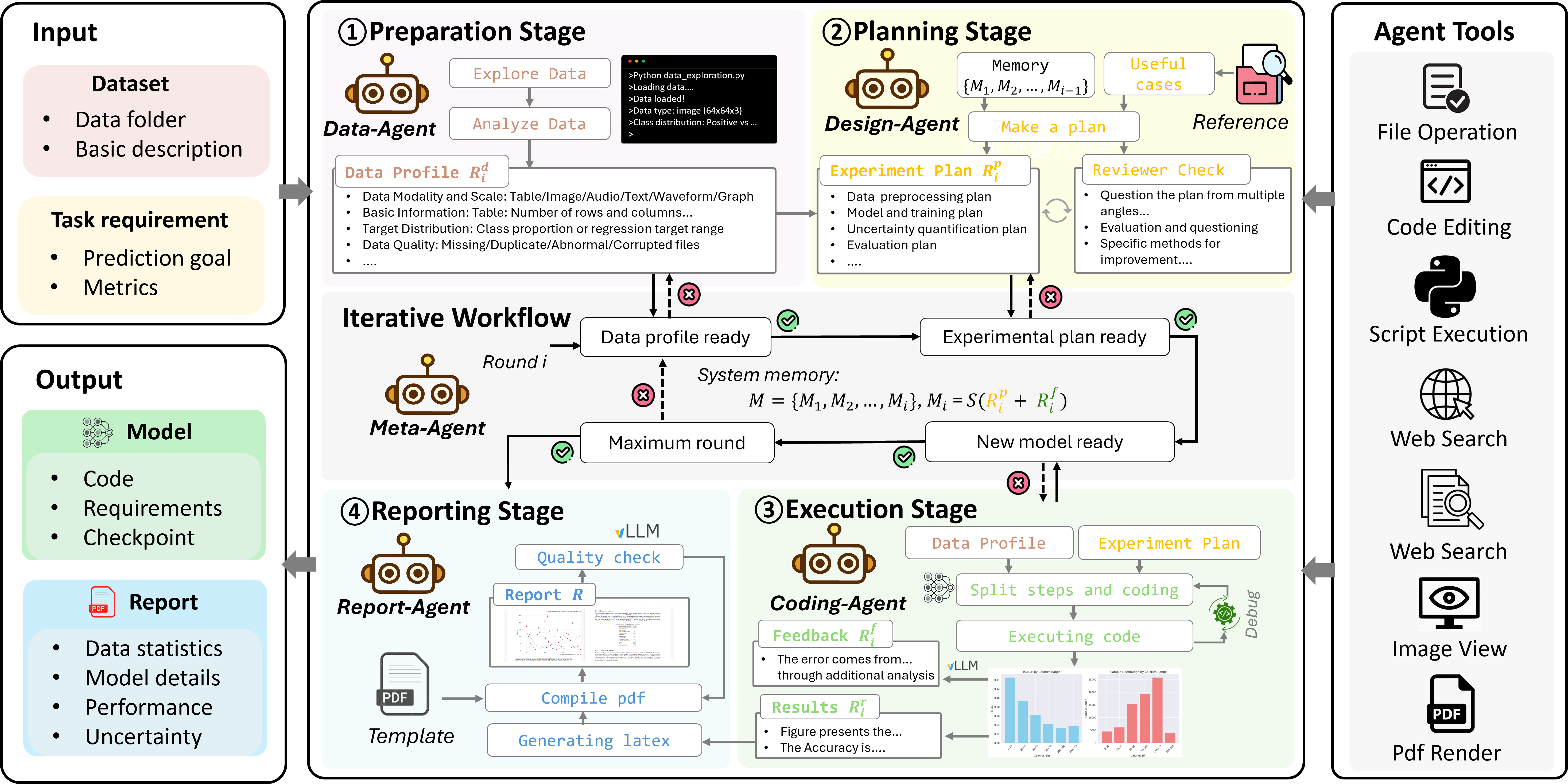}
    \caption{\textit{AutoHealth framework.} The system adopts a closed-loop, multi-agent architecture coordinated by a \textit{Meta-Agent}, which orchestrates an iterative workflow spanning preparation, planning, execution, and reporting. Specialized agents operate at each stage to enable modular, reliable, and reproducible health data modeling.}
    \label{fig:2}
     \vspace{-5mm}
\end{figure*}



\section{AutoHealth}

This section describes the proposed multi-agent system \textit{AutoHealth}, and its iterative workflow, as shown in Figure~\ref{fig:2}.

\subsection{Agent Specifications}

We provide brief descriptions of the agents in the framework below (detailed definitions can be found from Appendix~\ref{sec:prompt}).

\textbf{Meta-Agent} ($\mathcal{A}_m$) acts as the central coordinator and the primary interface with the user. It translates health data modeling objectives into global plans, orchestrates interactions among specialized agents, and manages an adaptive closed-loop feedback process to iteratively refine modeling decisions and track overall progress.

\textbf{Data-Agent} ($\mathcal{A}_d$) is responsible for health data exploration. It handles multi-source and heterogeneous inputs, performs data quality assessment, and produces structured data profiles to ensure that downstream stages operate on clinically meaningful and data-aware representations.

\textbf{Design-Agent} ($\mathcal{A}_p$) aims to generate an adaptive experimental plan under task-specific and uncertainty-aware constraints. Guided by modeling objectives and data characteristics, it proposes model architectures, feature configurations, and training strategies beyond rigid template-based reuse.

\textbf{Coding-Agent} ($\mathcal{A}_c$) translates high-level design specifications into executable and reproducible machine learning code. It implements data pipelines, model architectures, training procedures, and evaluation routines, while supporting iterative updates during closed-loop optimization.

\textbf{Report-Agent} ($\mathcal{A}_r$) synthesizes experimental decisions, results, and reliability analyses into structured reports. By summarizing model behavior, uncertainty, and limitations in an interpretable manner, it supports transparent, trustworthy, and reproducible use of the generated models.

 \begin{table*}[t]
\centering
\caption{Summary of inputs and outputs at each stage of the iterative workflow round $i$.}
\label{tab:2}
\resizebox{\textwidth}{!}{
\begin{tabular}{llll}
\hline
\textbf{Stage} & \textbf{Chief Agent} & \textbf{Input} & \textbf{Output} \\
\hline
Preparation 
& Data-Agent $\mathcal{A}_d$ 
& Raw data, task specification 
& Data profile $R_i^d$  \\

Planning 
& Design-Agent $\mathcal{A}_p$ 
& Data profile $R_i^d$, task specification,  memory $\{M_1,...,M_{i-1}\}$
& Experimental plan $R_i^p$  \\

Execution 
& Coding-Agent $\mathcal{A}_c$ 
& Raw data, data profile $R_i^d$, experimental plan $R_i^p$ 
& Experimental results $R_i^r$, experimental feedback $R_i^f$  \\

Reporting 
& Report-Agent $\mathcal{A}_r$ 
& Experimental results $R_i^r$  (only when $i=N$)
& Readability-aware report $R$\\
\hline
\end{tabular}
}
\vspace{-3mm}
\end{table*}
 
\subsection{Overall Iterative Workflow}

Under the coordination of the Meta-Agent $\mathcal{A}_m$, \textit{AutoHealth} operates an iterative, closed-loop workflow that progressively refines the modeling solution. 
Each iteration $i < N$ may involve \textit{up to three stages}: preparation, planning, and execution, which are selectively activated based on the current task state and intermediate outcomes, while the final reporting stage is triggered only after the last iteration $N$.

At a high level, the preparation stage focuses on data understanding, the planning stage translates validated data insights into task-conditioned experimental designs, and the execution stage implements and evaluates the proposed designs while generating feedback for refinement.

\begin{itemize}[leftmargin=1.2em, itemsep=0pt, topsep=0pt, parsep=0pt]

\item \textbf{Preparation Stage.}  
The Data-Agent performs systematic data exploration and quality assessment, producing a structured data profile $R_i^d$ that summarizes data modalities, distributions, temporal structures, and potential quality issues.

\item \textbf{Planning Stage.}  
Conditioned on the data profile $R_i^d$ and task specification, the Design-Agent generates an experimental plan $R_i^p$, including model architectures, training strategies, and evaluation protocols. For $i>1$, insights from previous rounds $\{M_1,\ldots,M_{i-1}\}$ are incorporated to support adaptive refinement.

\item \textbf{Execution Stage.}  
Given $R_i^d$ and $R_i^p$, the Coding-Agent implements the experimental plan and produces trained models together with an experimental record $R_i^r$. Structured feedback $R_i^f$ is generated based on model behavior and performance to guide subsequent iterations.

\end{itemize}

\textbf{Final Reporting Stage.}  
After the final iteration $N$, the Report-Agent consolidates $R_N^r$ into a reliability-aware, application-oriented report $R$, which serves as the final system output.

Table~\ref{tab:2} summarizes the inputs and outputs of each stage, and the following subsections describe their designs in detail.

\subsection{Preparation Stage}

Given the raw health data and task specification, the Data-Agent $\mathcal{A}_d$ produces a structured data profile $R_i^d$ to support downstream modeling. 
To achieve effective and systematic data understanding, we adopt three design principles.

\textbf{Code-driven Data Understanding.}
Rather than relying on user-provided dataset descriptions, $\mathcal{A}_d$ directly interacts with raw data through executable code, including file inspection, data loading, and programmatic analysis. 
This design assumes only minimal input from health experts (e.g., file paths and brief summaries) and enables detailed understanding of data structure and quality.

\textbf{Structured Objectives Across Data Types.}
Standardized exploration objectives are defined for seven common medical data types, including tabular, time-series, image, image segmentation, audio, free text, and graph data. 
For each type, the objectives cover basic properties, data quality indicators, and deeper characteristics such as feature--target relationships or lightweight feature screening.

\textbf{Data-informed Guidance for Downstream Modeling.}
Based on the exploration results, $\mathcal{A}_d$ generates reference code for data loading and provides preprocessing and modeling recommendations (e.g., handling class imbalance or selecting weighted losses), explicitly linking data characteristics to subsequent modeling decisions.

\subsection{Planning Stage}

Given the data profile $R_i^d$, task specification, and system memory from previous rounds $\{M_1,\ldots,M_{i-1}\}$, the Design-Agent $\mathcal{A}_p$ produces an experimental plan $R_i^p$. 
To improve executability and task suitability, three design principles are adopted.

\textbf{Retrieval-Augmented Planning.}
To avoid reliance on model prior knowledge alone, $\mathcal{A}_p$ is equipped with local case retrieval and web search tools, following prior work such as \textit{DS-Agent}~\cite{guo2024ds}. Retrieved solutions serve as references to ground planning decisions in validated practices.

\textbf{Feedback-Aware Customization.}
\textit{AutoHealth} maintains a persistent memory $\{M_1,\ldots,M_{i-1}\}$ that accumulates execution feedback and design effectiveness across iterations. 
During planning, this memory is queried to assess the suitability of reused solutions under current data and task conditions. 
When historical evidence suggests mismatch or degraded performance, $R_i^p$ is selectively revised, enabling cumulative cross-round optimization beyond template reuse.

\textbf{Iterative Plan Critique and Refinement.}
After an initial plan is generated, the Design-Agent $\mathcal{A}_p$ invokes an auxiliary LLM-based review tool to assess feasibility and potential risks. The resulting critiques are used to iteratively refine $R_i^p$ before execution.

\subsection{Execution Stage}

Given the data profile $R_i^d$ and experimental plan $R_i^p$, the Coding-Agent $\mathcal{A}_c$ implements the pipeline and produces an experimental record $R_i^r$ together with structured feedback $R_i^f$. 
To improve robustness, debuggability, and analytical depth, we adopt three design principles.

\textbf{Step-wise Plan Decomposition and Execution.}
Rather than generating a monolithic program, $\mathcal{A}_c$ decomposes $R_i^p$ into fine-grained steps that are implemented and validated sequentially. 
This step-wise execution reduces error propagation, improves fault localization, and stabilizes complex experimental pipelines.

\textbf{Comprehensive Performance and Uncertainty Analysis.}
Beyond a single target metric, $\mathcal{A}_c$ performs systematic evaluation of predictive performance, calibration, uncertainty behavior, and potential failure modes. 
Key statistics and visual diagnostics (e.g., learning curves and uncertainty plots) are logged in $R_i^r$ to support post-hoc analysis and reproducibility.

\textbf{VLLM-Assisted Reflective Analysis.}
To enhance result interpretation, a vision-language model (VLLM) is invoked \emph{as an auxiliary analysis tool} by $\mathcal{A}_c$, rather than as an independent agent. 
By jointly examining numerical results, visual diagnostics, and execution logs, this tool helps identify issues such as data leakage, overfitting, or misaligned objectives. 
The resulting feedback is summarized into $R_i^f$.

After each execution round $i$, the Meta-Agent $\mathcal{A}_m$  summarizes the experimental plan $R_i^p$ and execution feedback $R_i^f$ into a compact memory unit $M_i$. 
Formally, the system memory is updated as $\mathcal{M} = \{M_1, \ldots, M_{i-1}, M_i\}$, where $M_i = S(R_i^p, R_i^f)$,
and $S(\cdot)$ denotes a summarization operator that extracts salient design decisions, performance outcomes, failure cases, and actionable insights. 
This \textbf{updated memory} is carried forward to subsequent planning rounds, enabling cumulative, cross-round refinement of modeling strategies.

\subsection{Reporting Stage}

After the final iteration $N$, the Report-Agent $\mathcal{A}_r$ summarizes the execution results $R_N^r$ into a structured, application-oriented report $R_N$ to support reliable interpretation and practical use. 
Three design principles are adopted.

\textbf{Schema-Driven Reporting.}
The report is generated following a predefined schema and organized into three sections: data analysis, model performance, and uncertainty assessment. 
This standardized structure ensures comprehensive and reproducible documentation of modeling decisions, results, and limitations.

\textbf{Application-Oriented Interpretation.}
Rather than academic exposition, the report emphasizes operational guidance. 
Model performance and uncertainty behavior are explained in clear, instruction-oriented language, with explicit recommendations on when predictions can be trusted and when human review should be prioritized.

\textbf{Visual Quality Review.}
A VLLM-based reviewer is used as an auxiliary tool to check the clarity and consistency of figures, tables, and accompanying text. 
The resulting feedback is applied to refine the report before finalization.

\section{Results}

We evaluate the effectiveness of \textit{AutoHealth} by comparing it with state-of-the-art counterparts on heterogeneous health datasets. Results are introduced below.

  \begin{figure*}[t]
    \centering
    \includegraphics[width=0.8\textwidth]{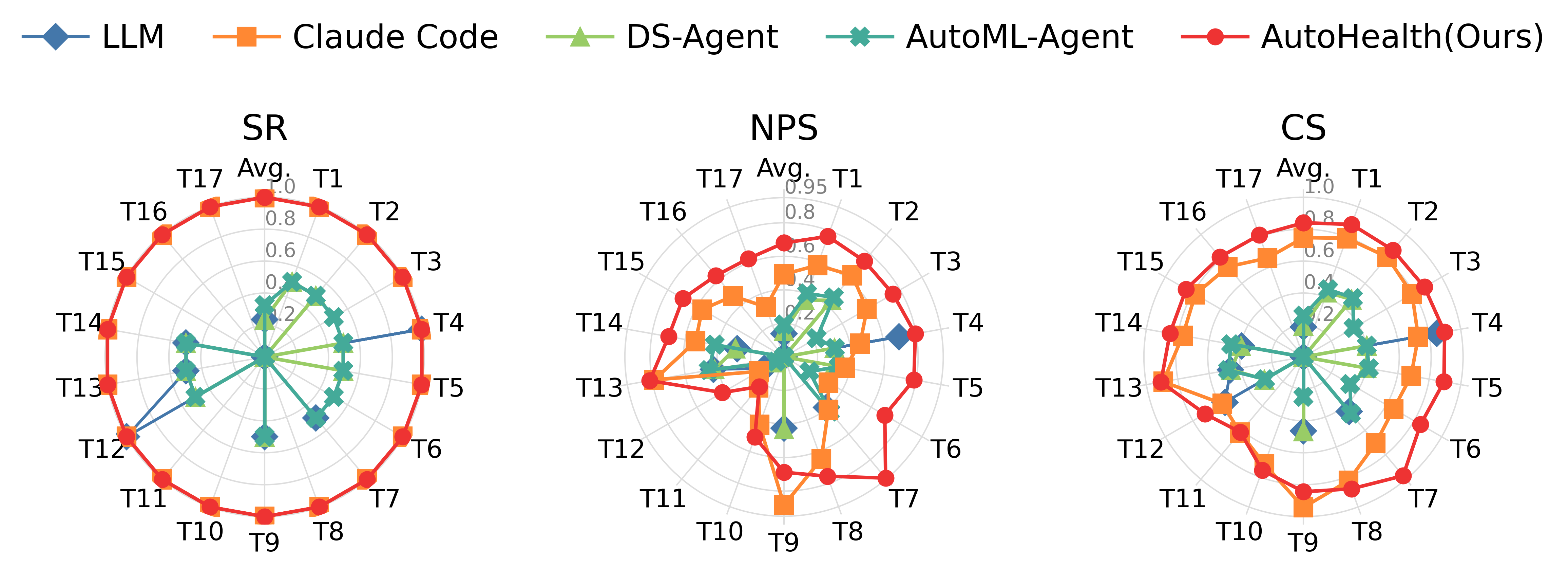}
    \caption{\textit{Performance comparison using the SR, NPS, and CS.} Avg presents the averaged score across all 17 tasks. }
    \label{fig:3}
\end{figure*}

\begin{figure*}[t]
    \centering
    \begin{subfigure}{0.3\linewidth}
        \includegraphics[width=\linewidth]{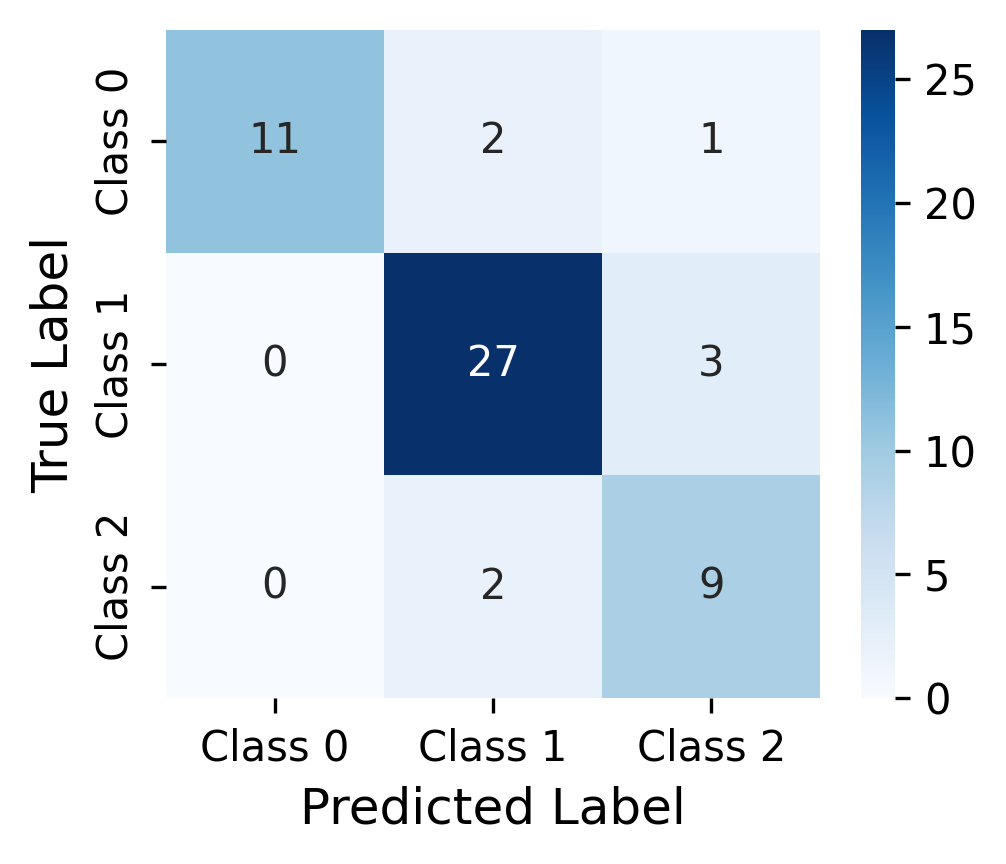}
        \caption{Confusion matrix}\label{fig:4a}
    \end{subfigure}
\hspace{0.0\linewidth}
    \begin{subfigure}{0.3\linewidth}
        \includegraphics[width=\linewidth]{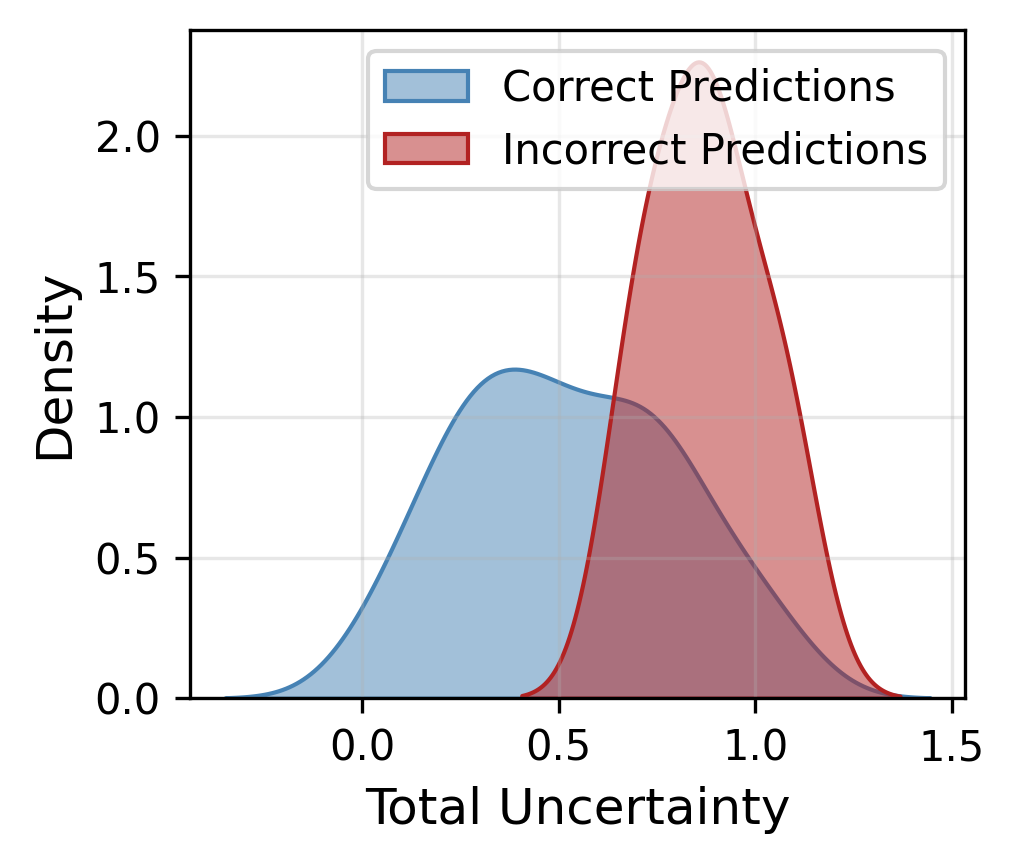}
        \caption{Uncertainty distribution}\label{fig:4b}
    \end{subfigure}
\hspace{0.0\linewidth}
    \begin{subfigure}{0.3\linewidth}
        \includegraphics[width=\linewidth]{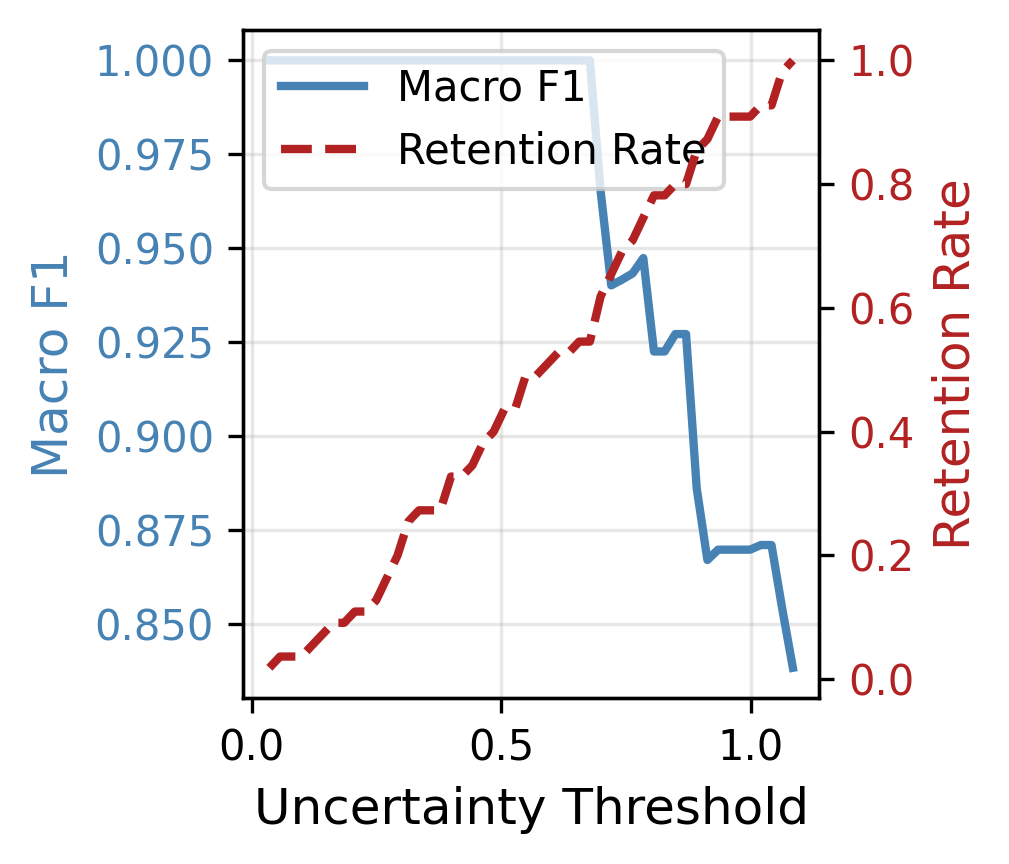}
        \caption{Selective prediction}\label{fig:4c}
    \end{subfigure}
   
    
    \caption{Excerpt of the system-generated report for Task 15 respiratory disease prediction. The complete report is provided in Appendix~\ref{sec:report}.}
    \label{fig:4}
\end{figure*}

\subsection{Setup}

\textbf{Benchmark.} Table~\ref{tab:3} summarizes our curated benchmark comprising 17 tasks. 
The datasets are drawn from heterogeneous sources, and task-specific evaluation metrics are adopted, such as accuracy for classification and root mean squared logarithmic error (RMSLE) for regression. 
Additional details on the datasets and evaluation metrics are provided in Appendix~\ref{sec:benchmark}. 
Compared with prior studies~\cite{guo2024ds,Trirat2025}, our benchmark covers a broader range of task types, posing a more stringent test of system generalizability.


\textbf{Evaluation Metrics.} 
Following prior work~\cite{Trirat2025}, we evaluate the system using \emph{Success Rate (SR)} and \emph{Normalized Performance Score (NPS)}, and report a \emph{Comprehensive Score (CS)} defined as $\text{CS}=0.5\times\text{SR}+0.5\times\text{NPS}$.

Specially, SR measures task completeness with two components: successful model training and uncertainty quantification. Completing both yields SR$=1$, completing only one yields SR$=0.5$, and completing neither yields SR$=0$.
NPS evaluates overall model quality by jointly considering task performance and uncertainty behavior.
For accuracy-based metrics, raw scores are used directly.
For loss-based metrics (termed as $s$), both performance and uncertainty metrics are transformed as
$S = 0.1 / (0.1 + s)$ to ensure scale consistency.
The transformed performance and uncertainty scores are then equally weighted to compute NPS.

\textbf{Baselines.} 
We compare our system with three categories of baselines. 
(1) \emph{naïve LLM}: an LLM is prompted with the same task specification, and the generated code is executed manually to record performance. 
(2) \emph{General coding agents}: we use \textit{Claude Code} with comparable prompts to ensure a fair comparison. Prompts used for \textit{LLM} and\textit{ Claude Code} can be found from Apeendix~\ref{sec:baseline}.
(3) \emph{Advanced data science agents}: we evaluate representative systems including \textit{DS-Agent} and \textit{AutoML-Agent}. 
We also attempted to include \textit{OpenLens AI}; however, its execution time was prohibitively long and all tasks exceeded practical runtime limits. We thus omit its quantitative results from the reported metrics.

\textbf{Implementation Details.} 
Unless otherwise specified, all agents are implemented using \textit{DeepSeek-V3.2} as the backbone model.  When VLLM is needed, we use \textit{GLM-4.6V}.
For all agent systems, including \textit{DS-Agent} and \textit{AutoML-Agent}, we set the maximum number of iteration rounds to $N=5$.
Each method is run independently three times, and the best performance is reported.
All experiments are conducted on an Ubuntu 22.04 LTS server equipped with four NVIDIA A100 GPUs.
To execute the generated models, we use the same runtime environment as \textit{DS-Agent}, with all required libraries pre-installed via the provided skeleton scripts.

\begin{table}[t]
\centering
\caption{Benchmark tasks across modalities.}
\label{tab:3}
\setlength{\tabcolsep}{3.5pt} 
\renewcommand{\arraystretch}{1.05} 
\resizebox{\linewidth}{!}{%
\begin{tabular}{lllll}
\toprule
\textbf{Modality} & \textbf{Task Type} & \textbf{ID} & \textbf{Name} & \textbf{Metric} \\
\midrule
\multirow{5}{*}{Tabular}
  & \multirow[t]{3}{*}{Classification}
    & 1  & Smoker status prediction                  & Accuracy \\
  &   & 2  & Patient survival prediction               & Accuracy \\
  &   & 3  & Drug response mechanism prediction        & Log Loss \\
  & Regression
    & 4  & Calorie estimation                        & RMSLE \\
  & Survival analysis
    & 5  & Post-HCT survival prediction              & C-index \\
\midrule
\multirow{3}{*}{Image}
  & \multirow[t]{2}{*}{Classification}
    & 6  & Skin cancer identification                & F1 \\
  &   & 7  & Melanoma detection                        & Accuracy \\
  & Segmentation
    & 8  & Retinal blood vessel segmentation         & Dice \\
\midrule
\multirow{4}{*}{Time series}
  & \multirow[t]{3}{*}{Classification}
    & 9  & ECG-based arrhythmia detection            & F1 \\
  &   & 10 & IMU-based gesture prediction              & Average F1 \\
  &   & 11 & EEG-based seizure prediction              & KL divergence \\
  & Forecasting
    & 12 & COVID-19 infection case forecasting       & MAE \\
\midrule
\multirow{2}{*}{Free text}
  & \multirow[t]{2}{*}{Classification}
    & 13 & Autism spectrum disorder prediction & Accuracy \\
  &   & 14 & COVID-19 sentiment classification         & Accuracy \\
\midrule
\multirow{2}{*}{Audio}
  & \multirow[t]{2}{*}{Classification}
    & 15 & Respiratory disease prediction            & Macro-F1 \\
  &   & 16 & Dysarthria detection                      & Accuracy \\
\midrule
Graph
  & Link prediction
    & 17 & Protein link property prediction          & Hits@20 \\
\bottomrule
\end{tabular}%
}
\vspace{-3mm}
\end{table}

\subsection{Main Results}
Overall results are summarized in Figure~\ref{fig:3}, with detailed metrics reported in Appendix~\ref{sec:results_2} Tables~\ref{tab:SR1}–\ref{tab:CS}.

\textbf{Success Rate.}
Figure~\ref{fig:3} shows that only \textit{Claude Code} and our proposed \textit{AutoHealth} successfully complete all tasks, whereas the other methods fail during either model construction or uncertainty quantification.
The naïve LLM baseline achieves reasonable performance only with carefully designed prompts and manual code execution.
\textit{DS-Agent} and \textit{AutoML-Agent} fail on several tasks (e.g., Tasks 10, 11, and 15–17), mainly due to limited modality generalization of their predefined pipelines.
Consequently, the average success rates (SR) for the naïve LLM, \textit{Claude Code}, \textit{DS-Agent}, \textit{AutoML-Agent}, and \textit{AutoHealth} are 23.5\%, \textbf{100\%}, 23.5\%, 32.3\%, and \textbf{100\%}, respectively.

\textbf{Performance Score.}
\textit{AutoHealth} achieves an average NPS of 0.667, exceeding the second-best \textit{Claude Code} of 0.493 by \textbf{37.9\%}. Specifically, its average prediction score is 0.747 (Table~\ref{tab:performance_metrics2}) and  uncertainty score is 0.613 (Table~\ref{tab:uncertainty_final2}), outperforming \textit{Claude Code} by \textbf{29.2\%}, \textbf{50.2\%}, respectively. 
Moreover, as shown in Figure~\ref{fig:3}, \textit{AutoHealth} attains state-of-the-art performance on 16 out of 17 tasks.


\textbf{Comprehensive Score.}
As shown in Figure~\ref{fig:3}, \textit{AutoHealth} achieves the highest average  CS of 0.840, outperforming \textit{Claude Code} (0.747), \textit{AutoML-Agent} (0.258), \textit{DS-Agent} (0.195), and the naïve LLM (0.187), corresponding to relative improvements ranging from \textbf{12.4\% to 349.2\%}.
These results demonstrate that \textit{AutoHealth} provides a reliable and effective end-to-end solution for automated health data modeling with minimal human intervention.

\subsection{Report Quality Assessment}
To illustrate \textit{AutoHealth}'s capability in generating high-quality, reliability-aware reports, we present a representative case study based on Task~15. This task focuses on respiratory disease prediction, aiming to distinguish healthy, diseased, and chronic conditions using multi-modal inputs. Due to its multi-modal structure and severe class imbalance, Task~15 is selected as a representative example. The complete automatically generated report is provided in Appendix~\ref{sec:report}, with selected figures shown in Figure~\ref{fig:4}.

This report demonstrates how \textit{AutoHealth} autonomously completes the full modeling pipeline on a real-world clinical dataset that integrates structured phenotypic variables (demographics, symptoms, and medical history) with respiratory audio recordings (cough and vowel sounds). Without human intervention, \textit{AutoHealth} selects a dual-branch late-fusion neural architecture to separately model tabular features and audio embeddings, followed by feature-level fusion for joint prediction. To improve robustness and support uncertainty estimation, a deep ensemble of three models is further constructed.

During training, \textit{AutoHealth} automatically identifies class imbalance and applies targeted mitigation strategies (i.e., class-weighted loss), resulting in strong and well-balanced performance with a test Macro-F1 score of 0.838 and similar recall across classes (Figure~\ref{fig:4a}).

In addition, the automatically generated report includes calibration and uncertainty analyses, showing that misclassified samples are consistently associated with higher predictive uncertainty (Figure~\ref{fig:4b}). This supports effective separation of high- and low-confidence predictions and enables risk-aware, human-in-the-loop deployment. As quantified in Figure~\ref{fig:4c}, rejecting the 20\% most uncertain samples increases the Macro-F1 on the remaining predictions to 0.928, at the cost of abstaining from some cases.

Overall, \textit{AutoHealth} generates not only accurate predictions but also actionable diagnostic insights and deployment guidance by explicitly identifying failure modes and leveraging uncertainty for risk-aware clinical decision-making.

 \subsection{Additional Analysis}

\textbf{Ablation Study.} We first examine the contributions of data exploration and step-wise plan decomposition and execution to successfully completing all tasks. We hypothesize that these design components are key to generalizing across heterogeneous data modalities and learning settings. We compare the SR of the full AutoHealth system with two degraded variants: \textit{AutoHealth-S} (without step decomposition) and \textit{AutoHealth-SD} (without both step decomposition and data exploration). The results are summarized in Table~\ref{tab:ablation_sr}. Removing step decomposition already results in a notable reduction in success rate, while further disabling data exploration leads to an even more pronounced performance drop. This trend underscores the complementary importance of structured planning and data-aware reasoning in \textit{AutoHealth}.

\begin{table}[t]
\centering
\caption{Ablation results on success rate (SR).}
\label{tab:ablation_sr}
\scriptsize
\begin{tabular}{lccc}
\toprule
Method & AutoHealth & AutoHealth-S & AutoHealth-SD \\
\midrule
SR (\%) & 100 & 70.6 & 41.2 \\
\bottomrule
\end{tabular}
\vspace{-2mm}
\end{table}

\begin{table}[t]
\centering
\caption{Ablation results on uncertainty quantification. AVG denoted the average normalized uncertainty score across task 1 to 5 as detailed in Appendix Table~\ref{tab:abalation2}.}
\label{tab:ablation_un}
\scriptsize
\begin{tabular}{lcccc}
\toprule
Method & Claude Code & AutoHealth & AutoHealth-T & AutoHealth-R \\
\midrule
AVG  &   0.403 & 0.819 & 0.733 & 0.650 \\
\bottomrule
\end{tabular}
\vspace{-5mm}
\end{table}

We then examine the trade-off between template-based solution reuse and objective-aware adaptation by ablating uncertainty quantification templates during planning (\textit{AutoHealth-T}) and VLLM-assisted reflective analysis during execution (\textit{AutoHealth-R}). We evaluate these variants on tabular tasks (T1–T5), where predictive performance is largely comparable across methods, enabling a focused analysis of uncertainty estimation behavior. As shown in Table~\ref{tab:ablation_un}, the full \textit{AutoHealth} achieves substantially lower uncertainty error, whereas removing either component results in consistent performance degradation. Specifically, compared to \textit{AutoHealth}, the normalized uncertainty score decreases by 10.5\% for \textit{AutoHealth-T} and by 20.6\% for \textit{AutoHealth-R}. These results indicate that reliable uncertainty estimation requires both reusable templates and adaptive, feedback-driven reasoning, rather than template reuse alone.

\begin{table}[t]
\centering
\caption{Token and time cost for part of tasks. Input, output and cache token counts are reported in millions (M).}
\label{tab:token_cost}
\scriptsize
\begin{tabular}{llccccc}
\toprule
\textbf{Task} & \textbf{Data size} &
{\textbf{Input}} &
{\textbf{Output}} &
{\textbf{Cache}} &
{\textbf{Cost(rmb)}} &
{\textbf{Time(min)}} \\
\midrule
T1  & 11.03MB  & 1.58 & 0.20 & 0.93 & 2.1 & 119.8 \\
T5  & 9.87MB   & 2.65 & 0.38 & 1.49 & 3.8 & 246.1 \\
T6  & 143.5MB  & 3.49 & 0.55 & 1.76 & 5.5 & 478.7 \\
T9  & 7.75MB   & 1.81 & 0.22 & 1.14 & 2.2 & 123.2 \\
T11 & 6.6GB    & 3.73 & 0.47 & 1.83 & 5.6 & 634.1 \\
T15 & 1.9GB    & 3.03 & 0.38 & 1.69 & 4.2 & 187.5 \\
T17 & 102MB    & 3.38 & 0.47 & 1.77 & 5.0 & 419.5 \\
\bottomrule
\end{tabular}
\vspace{-8mm}
\end{table}

\textbf{Cost and Efficiency Analysis.} 
Table~\ref{tab:token_cost} presents the token usage, monetary cost, and wall-clock time for some tasks (full results can be found from Appendix Table~\ref{tab:token_cost_2} and execution time proportion of each stage is summarized in Appendix Figure~\ref{fig:time}). Across the 17 tasks, the total token consumption remains within a moderate range, while the monetary cost is consistently low, typically below 6 rmb per task.
Importantly, \textit{AutoHealth} demonstrates fast iterative cycles, with most tasks completing within 2–6 hours (5 runs), even for tasks involving gigabyte-scale data. This efficiency is largely attributed to effective and structured multi-agent coordination, which substantially reduce redundant token consumption across iterations. By integrating task understanding, model construction, execution, and feedback into a unified end-to-end agentic workflow, \textit{AutoHealth} minimizes human intervention and iteration overhead. Together, these results indicate that \textit{AutoHealth} enables rapid and affordable end-to-end experimentation, thereby lowering the barrier to large-scale empirical studies and accelerating scientific discovery in real-world health data modeling.
\section{Conclusions}

In this work, we presented \textit{AutoHealth}, a novel uncertainty-aware, closed-loop multi-agent framework for autonomous health data modeling. By coordinating specialized agents for data exploration, task-conditioned model construction, optimization, and reporting, \textit{AutoHealth} addresses key limitations of existing LLM-based data science agents in handling heterogeneous health data and reliability-critical requirements. Extensive experiments on a challenging real-world benchmark demonstrate that \textit{AutoHealth} consistently outperforms strong general-purpose baselines across diverse modalities and learning settings, while providing uncertainty-aware outputs to support trustworthy interpretation. Overall, this work advances autonomous data science toward more reliable and practical deployment in safety-critical settings like healthcare.




\section*{Software and Data}
Our code has been submitted as to anonymous Github. All data used for this study are publicly available.

\section*{Impact Statement}
This paper presents a methodological contribution aimed at advancing autonomous data science and machine learning systems for health data modeling. The proposed framework is designed to improve reliability, transparency, and uncertainty awareness in predictive modeling, which are widely recognized as important considerations for healthcare applications. While such systems may support more efficient model development and analysis, they are not intended to replace clinical expertise or decision-making. As with other machine learning approaches applied to health data, appropriate data governance, validation, and human oversight remain essential to ensure responsible use. We do not foresee any immediate negative societal impacts beyond those commonly associated with machine learning research in healthcare.

\bibliography{ref}
\bibliographystyle{icml2026}

\newpage
\appendix
\onecolumn

 \section*{Appendix}

The appendix of this paper is organized as follows:
\begin{itemize}
    
    \item Appendix~\ref{sec:A1} provides additional details on the benchmark.
    \item Appendix~\ref{sec:results_2} presents more detailed results.
    \item Appendix~\ref{sec:report} presents the generated report for task 15.
    \item Appendix~\ref{sec:prompt} specifies the design of the specialized agents in \textit{AutoHealth}.
\end{itemize}

\section{Additional Introduction for Benchmark and Experiments}\label{sec:A1}

\subsection{Benchmark}\label{sec:benchmark}

Tabular data based tasks include: 

\begin{itemize}
    \item \textbf{T1 Smoker status prediction.}  This task is built on a structured health‑behavior dataset where each row represents an individual’s physiological signals and lifestyle attributes. The target is a binary label indicating smoking status. The goal is to learn patterns in biomarkers—such as lung capacity, cholesterol levels, blood pressure, or other body‑signal indicators—that reliably distinguish smokers from non‑smokers. For ML, the dataset offers opportunities to explore feature importance, uncover latent health correlations, evaluate model fairness across subgroups, and build predictive systems that highlight which physiological signals are most informative for identifying smoking‑related risk profiles. More details can be found from \url{https://www.kaggle.com/datasets/kukuroo3/body-signal-of-smoking/data}.

    \item \textbf{T2 Patient survival prediction.} This task centers on a structured clinical dataset where each record captures a patient’s demographic attributes, laboratory measurements, and disease‑related indicators, and the target is a categorical survival outcome reflecting different stages of prognosis. The aim is to model how patterns in these medical features relate to survival trajectories, enabling ML systems to learn which biomarkers and clinical signals are most predictive of patient risk and long‑term outcomes.  More details can be found from \url{https://www.kaggle.com/datasets/joebeachcapital/cirrhosis-patient-survival-prediction/code?datasetId=3873965&sortBy=voteCount}.

    \item \textbf{T3 Drug response mechanism prediction.} This task uses a high‑dimensional biological screening dataset where each sample corresponds to a compound tested under specific cellular conditions, and the targets are multi‑label mechanism‑of‑action indicators describing which biological pathways the compound perturbs. The goal is to learn predictive mappings from gene‑expression and cell‑viability features to these MoA labels, enabling models to uncover functional signatures, distinguish subtle pathway effects, and support downstream drug‑discovery insights. More details can be found from \url{https://www.kaggle.com/competitions/lish-moa/data?select=train_targets_nonscored.csv}.

    \item \textbf{T4 Calorie estimation.} This task provides a structured regression task built on \textit{train.csv}, where each row represents a physical‑activity record with features such as duration, heart‑rate signals, and movement‑related metrics, and the target is a continuous Calories value that quantifies energy expenditure . The goal is to learn how patterns in these activity features map to calorie burn, enabling models to capture physiological relationships, improve prediction accuracy, and support insights into how different forms of movement translate into metabolic output. More details can be found from \url{https://www.kaggle.com/competitions/international-medical-ai-and-data-science-calorie/data}.

    \item \textbf{T5 Post-HCT survival analysis.} This task is built on a synthetic but clinically grounded dataset where each row represents an allogeneic HCT patient with demographic variables, disease characteristics, transplant details, and post‑treatment clinical signals, and the target is a real‑valued risk score reflecting predicted survival likelihood after transplant. The challenge is to model survival outcomes in a way that is both accurate and equitable across racial groups, learning patterns in the medical features that drive post‑HCT prognosis while minimizing disparities in predictive performance. More details can be found from \url{https://www.kaggle.com/competitions/equity-post-HCT-survival-predictions/overview}.

\end{itemize}

Clinical image data based tasks include: 

\begin{itemize}
     \item \textbf{T6 Skin cancer identification .} This task provides a skin‑lesion image dataset where part of the training set is labeled and a substantial portion is unlabeled, and the labeled subset is heavily imbalanced with far more benign than malignant cases. The task is to build an image‑classification model that predicts a probability for the LABEL variable (malignant vs. benign) for each test image, learning effectively from limited labeled data while leveraging the large unlabeled pool and mitigating class imbalance to achieve strong F1 performance. More details can be found from \url{https://www.kaggle.com/competitions/dsm-2024-competition/overview}.
 
    \item \textbf{T7 Melanoma detection.} This task is built around a curated skin‑lesion image dataset in which each sample is a dermoscopic photograph paired with a binary target label indicating whether the lesion is benign or malignant. The goal is to train a classifier that learns visual patterns—such as texture, color variation, and structural irregularities—that distinguish malignant melanoma from non‑cancerous lesions, enabling accurate probability predictions for the target label on unseen images. More details can be found from \url{https://www.kaggle.com/datasets/bhaveshmittal/melanoma-cancer-dataset}.

    \item \textbf{T8 Retinal blood vessel segmentation.} This task gives a retinal fundus image dataset in which each sample is a high‑resolution photograph of the eye, paired with a pixel‑level segmentation mask marking blood‑vessel locations as 1 and background as 0. The goal is to train a model that learns the visual structure of retinal vasculature—capturing vessel width, branching patterns, and subtle pathological variations—to accurately predict a full‑resolution vessel mask for each input image, supporting downstream diagnostic and ophthalmic analysis. More details can be found from \url{https://www.kaggle.com/datasets/abdallahwagih/retina-blood-vessel/data}.
    
\end{itemize}

Health time series data based tasks include: 

\begin{itemize}
     \item \textbf{T9 ECG-based arrhythmia detection.} This task uses a structured ECG time‑series dataset where each row corresponds to a single patient and contains a heartbeat of ECG signal value along with metadata such as an ID and rhythm‑class indicators. The target is the healthy status label, a binary outcome indicating whether the ECG is assessed as Normal or Abnormal, derived from clinical annotations in the original PhysioNet source. The goal is to learn discriminative patterns in the raw ECG waveform—morphology, rhythm regularity, and subtle signal deviations—to accurately classify cardiac health status for new, unseen patients. More details can be found from \url{https://www.kaggle.com/competitions/ecg-ai-fast-checkup/overview}.
 
    \item \textbf{T10 IMU-based gesture prediction.} This task provides a multichannel time‑series sensor dataset collected from a wrist‑worn device, where each sequence contains synchronized IMU motion signals along with temperature and proximity readings. The target is a gesture label indicating whether the movement corresponds to a body‑focused repetitive behavior (BFRB) or a non‑BFRB action, with further granularity for specific BFRB gesture types. The task is to train a model that learns discriminative temporal patterns across sensors to reliably classify these behaviors, supporting more accurate and clinically meaningful detection of BFRB‑related movements. Binary-F1 on whether the gesture is one of the target or non-target types, and Macro-F1 on gesture, where all non-target sequences are collapsed into a single non-target class, are calculated. The final score is the average of the binary F1 and the macro F1 scores. More details can be found from \url{https://www.kaggle.com/competitions/cmi-detect-behavior-with-sensor-data}.

    \item \textbf{T11 EEG-based seizure prediction.} This task provides a large‑scale EEG dataset where each training example corresponds to a short segment of multichannel brain‑wave activity recorded from critically ill patients, paired with expert‑annotated vote counts for six possible harmful brain‑activity patterns: seizure, LPD, GPD, LRDA, GRDA, and other. The task is to learn a model that maps raw EEG signals and derived features to a probability distribution over these six target classes, capturing subtle temporal–spectral signatures so the system can reliably classify harmful patterns and support faster, more consistent neurocritical‑care assessment More details can be found from \url{https://www.kaggle.com/competitions/hms-harmful-brain-activity-classification/overview}.

    \item \textbf{T12 COVID-19 infection case forecasting.} This dataset provides a global COVID‑19 time‑series table where each row records a specific country or province on a given date, along with key epidemiological counts: Confirmed, Recovered, and Deaths, supported by geographic metadata such as latitude and longitude. The task is to model how these case trajectories evolve over time—treating the target as any of these outcome variables depending on the analysis goal—so that ML systems can capture temporal patterns, forecast future case numbers, and uncover relationships between regional characteristics and pandemic dynamics. More details can be found from \url{https://www.kaggle.com/datasets/rahulgupta21/datahub-covid19/data}.
    
\end{itemize}

Free text data based tasks include: 

\begin{itemize}
     \item \textbf{T13 Autism spectrum disorder prediction.} This task is built on the TASD dataset, where each sample is a short textual description of a toddler’s daily behavior annotated with structured behavioral features such as attention response, word repetition, emotional empathy, noise sensitivity, and more, all derived from early ASD assessment cues. The target is a binary ASD label indicating whether the described behavior corresponds to an ASD or non‑ASD toddler. The goal is to learn linguistic and semantic patterns in these behavior‑focused narratives so that ML models can detect early ASD‑related signals from natural text, supporting more scalable and interpretable early‑screening tools. More details can be found from \url{https://www.kaggle.com/datasets/sharafatahmed/tasd-dataset-text-based-early-asd-detection/data}.
 
    \item \textbf{T14 COVID-19 sentiment classification.} This dataset contains thousands of COVID‑19–related tweets, each record providing the tweet text along with metadata such as location and timestamp, and the target is a sentiment label manually assigned to reflect the tweet’s emotional tone (e.g., positive, negative, neutral, extremely positive, extremely negative) . The task is to train a text‑classification model that learns linguistic and contextual cues in the tweets to accurately predict sentiment, enabling analysis of public mood, misinformation patterns, and real‑time reactions during the pandemic. More details can be found from \url{https://www.kaggle.com/datasets/datatattle/covid-19-nlp-text-classification/data}.
 
\end{itemize}

Audio data based tasks include: 

\begin{itemize}
     \item \textbf{T15 Respiratory disease prediction.} The goal is to build a multiclass classifier that predicts a disease label for each individual—distinguishing healthy, diseased, and chronic respiratory conditions using the provided audio and metadata. Performance is evaluated using macro F1‑score, reflecting the dataset’s class imbalance and the need for balanced sensitivity across all disease categories. More details can be found from \url{https://www.kaggle.com/competitions/airs-ai-in-respiratory-sounds/overview}.
 
    \item \textbf{T16 Dysarthria detection.} The task focuses on detecting whether a speech recording comes from a speaker with dysarthria. The data consists of short audio clips paired with minimal speaker information, capturing natural variations in articulation, rhythm, and vocal quality. The model’s goal is to learn patterns in these acoustic signals that reliably separate dysarthric from typical speech, without relying on any dataset‑specific file structures or formatting details. More details can be found from \url{https://www.kaggle.com/datasets/iamhungundji/dysarthria-detection/data}.
 
\end{itemize}

Graph data based tasks include: 
\begin{itemize}
     \item \textbf{T17 Protein link property prediction.} The ogbl‑ddi dataset from the Open Graph Benchmark is designed for link prediction in a large‑scale biological interaction graph. It captures a network where nodes represent drugs, and edges represent known drug–drug interactions curated from biomedical knowledge bases. The machine learning task focuses on predicting whether a previously unseen pair of drugs should form an interaction edge, based on the structural patterns of the graph. This makes it a clean, well‑defined benchmark for evaluating graph neural networks on real‑world pharmacological interaction discovery, with direct relevance to drug safety, polypharmacy risk modeling, and computational drug repurposing. More details can be found from \url{https://ogb.stanford.edu/docs/linkprop/}.
 
\end{itemize}

\newpage
A subset of representative task specifications and basic data descriptions for the LLM or agent systems is presented below.

\begin{PromptBox}{Task 1}
**Task Name: Smoker Status Prediction**

**[Task Overview]**
- Task Type: tabular binary classification
- Modality: structured tabular data (health check / bio-signal related features)
- Downstream Goal: predict whether a person is a smoker based on clinical measurements and basic health indicators.
- Target Column: smoking (0 = non-smoker, 1 = smoker)
- Evaluation Metric: Accuracy
- Objective: train a high-performing model that generalizes well to unseen individuals for this classification task.

**[Dataset Paths]**
- Dataset Root: /root/Dataset/classifying_smoker_status/
- Train: /root/Dataset/classifying_smoker_status/data/train/train.csv
- Validation: /root/Dataset/classifying_smoker_status/data/val/val.csv
- Test: /root/Dataset/classifying_smoker_status/data/test/test.csv

**[Dataset Details]**
- Format: CSV
- Columns (brief):
  - ID: sample identifier
  - gender: gender
  - age: age
  - height(cm), weight(kg), waist(cm): basic body measurements
  - eyesight(left/right): eyesight measurements
  - hearing(left/right): hearing measurements
  - systolic, relaxation: blood pressure related measurements
  - fasting blood sugar: fasting blood glucose
  - Cholesterol, triglyceride, HDL, LDL: lipid profile indicators
  - hemoglobin: hemoglobin level
  - Urine protein: urine protein indicator
  - serum creatinine: serum creatinine
  - AST, ALT, Gtp: liver-related indicators
  - oral: oral examination status
  - dental caries: dental caries indicator
  - tartar: tartar status
  - smoking: target label
\end{PromptBox}

\begin{PromptBox}{Task 8}
**Task Name: Retinal Blood Vessel Segmentation**

**[Task Overview]**
- Task Type: medical image segmentation (binary segmentation)
- Modality: retinal fundus images with pixel-level vessel masks
- Downstream Goal: segment retinal blood vessels from a fundus image.
- Target: binary mask image (vessel pixels vs background pixels)
- Evaluation Metric: Dice Coefficient
- Objective: train a high-performing segmentation model that produces accurate vessel masks on unseen retinal images.

**[Dataset Paths]**
- Dataset Root: /root/Dataset/Retina_Blood_Vessel_for_segmentation/
- Train Images: /root/Dataset/Retina_Blood_Vessel_for_segmentation/data/train/image/
- Train Masks: /root/Dataset/Retina_Blood_Vessel_for_segmentation/data/train/mask/
- Validation Images: /root/Dataset/Retina_Blood_Vessel_for_segmentation/data/val/image/
- Validation Masks: /root/Dataset/Retina_Blood_Vessel_for_segmentation/data/val/mask/
- Test Images: /root/Dataset/Retina_Blood_Vessel_for_segmentation/data/test/image/
- Test Masks: /root/Dataset/Retina_Blood_Vessel_for_segmentation/data/test/mask/

**[Dataset Details]**
- Format: PNG images
- Paired files (brief):
  - image/: retinal fundus images
  - mask/: corresponding binary vessel masks with the same filename
\end{PromptBox}

\begin{PromptBox}{Task 15}
**Task Name: Respiratory Disease Prediction**

[**Task Overview**]
- Task Type: multi-modal classification (tabular + audio)
- Modality: structured tabular metadata + respiratory audio recordings (cough and vowel)
- Downstream Goal: predict the respiratory disease class for each candidate based on metadata and audio signals.
- Target Column: disease (categorical label)
- Evaluation Metric: Macro-F1
- Objective: train a high-performing model that generalizes well to unseen candidates for robust respiratory disease detection.

[**Dataset Paths**]
- Dataset Root: /root/Dataset/medical_sound_classification/
- Train Table: /root/Dataset/medical_sound_classification/data/train/train.csv
- Train Sounds: /root/Dataset/medical_sound_classification/data/train/sounds/
- Validation Table: /root/Dataset/medical_sound_classification/data/val/val.csv
- Validation Sounds: /root/Dataset/medical_sound_classification/data/val/sounds/
- Test Table: /root/Dataset/medical_sound_classification/data/test/test.csv
- Test Sounds: /root/Dataset/medical_sound_classification/data/test/sounds/

[**Dataset Details**]
- Format: CSV (tables) + WAV (audio) + JSON (precomputed embeddings)
- Table columns (brief):
  - candidateID: candidate identifier (maps to the corresponding audio folder)
  - age, gender: basic demographics
  - tbContactHistory, wheezingHistory, phlegmCough, familyAsthmaHistory, feverHistory, coldPresent: symptom / history indicators
  - packYears: smoking exposure indicator
  - disease: target label
- Audio/embedding files (brief):
  - Each candidate folder contains cough.wav, vowel.wav, and optional embedding files (emb_cough.json, emb_vowel.json).
\end{PromptBox}

\newpage
\subsection{Baseline}\label{sec:baseline}

\begin{PromptBox}{Prompt for LLM}
You are a senior machine learning engineer. Your objective is to build a model with optimal predictive performance and to select an appropriate uncertainty estimation method to integrate with the model, such that uncertainty-related metrics (e.g., ECE, Brier Score, PCIP) can be computed and reported. Based on the user's **task description**, write a complete Python script.

Requirements:
1. The script must load the data directly from the provided file paths.
2. The script must compute and print all evaluation metrics specified by the task.
3. The code must be enclosed within python and blocks.
4. Do not use the input() function; the script must run fully automatically.

In addition to completing the modeling task, select an appropriate uncertainty estimation method and output the corresponding uncertainty analysis results.
\end{PromptBox}

\begin{PromptBox}{Prompt for Claude Code}
Please read the task file **task description** carefully and strictly follow all instructions without asking for clarification.

Your task is to train a model that achieves optimal predictive performance and to select an appropriate uncertainty estimation method to integrate with the model, such that uncertainty-related metrics (e.g., ECE, Brier Score,) can be computed and reported.

Before starting each task, create a corresponding task directory under 'OUTPUTROOT', named using the task name and a timestamp.

All outputs and execution scripts for each task must be placed in the corresponding directory. After completion, provide a brief summary of the executed steps and results, and must report both model performance metrics and uncertainty analysis results. If the task fails, explicitly report the failure. The final output should be written to a file named result.txt.

Do not ask for confirmation. Directly execute file writing and command execution. All code should be run directly by you.

\end{PromptBox}

\section{Additional Details for Results}\label{sec:results_2}

This section presents the detailed experimental results.  Tables~\ref{tab:performance_metrics2} and \ref{tab:uncertainty_final2} report normalized downstream performance for direct comparison.

\textbf{Task Implementation} Figure~\ref{fig:3} shows that only \textit{Claude Code} and our proposed \textit{AutoHealth} are able to successfully complete all tasks end to end, whereas the other methods fail either during model construction or uncertainty quantification. The naïve LLM baseline can achieve reasonable performance only when carefully designed prompts are provided and the generated code is executed manually.
\textit{DS-Agent} and \textit{AutoML-Agent} fail on several tasks (e.g., Tasks 10, 11, 15, 16, and 17), primarily due to the limited modality generalization of their predefined pipelines. Overall, across all tasks, the model success rates for the naïve LLM, \textit{Claude Code}, \textit{DS-Agent}, \textit{AutoML-Agent}, and \textit{AutoHealth} are 35.3\%, 100\%, 47.1\%, 64.7\%, and 100\% (Table~\ref{tab:SR1}), respectively.
In terms of uncertainty quantification, the corresponding success rates are 11.8\%, 100\%, 0\%, 0\%, and 100\% (Table~\ref{tab:SR2}).

\textbf{Downstream Performance} Tables~\ref{tab:performance_metrics2} and \ref{tab:uncertainty_final2} report the normalized downstream performance for direct comparison across methods. Our model achieves an average performance score of 0.731 and an average uncertainty score of 0.602, exceeding the second-best results (0.578 and 0.408, respectively) by 26.5\% and 47.5\%. As summarized in Figure~\ref{fig:3}, \textit{AutoHealth} attains state-of-the-art performance on 16 out of 17 tasks, while the remaining two tasks achieve results that are very close to the best-performing baselines.

\begin{table}[H] 
  \centering
  \caption{Comparison of code execution success rates. A value of 1 indicates success and 0 indicates failure.}
  \label{tab:SR1}
\scriptsize
    \begin{tabular}{lcccccccccccccccccc}
      \toprule
      Method & T1 & T2 & T3 & T4 & T5 & T6 & T7 & T8 & T9 & T10 & T11 & T12 & T13 & T14 & T15 & T16 & T17 &Pass rate\\
      \midrule
      LLM           & 0 & 0 & 0 & 1 & 0 & 0 & 1 & 0 & 1 & 0 & 0 & 1 & 1 & 1 & 0 & 0 & 0 & 35.3\%\\
      Claude Code   & 1 & 1 & 1 & 1 & 1 & 1 & 1 & 1 & 1 & 1 & 1 & 1 & 1 & 1 & 1 & 1 & 1 & 100\%\\
      DS-Agent     & 1 & 1 & 0 & 1 & 1 & 0 & 0 & 0 & 1 & 0 & 0 & 1 & 1 & 1 & 0 & 0 & 0  & 47.1\%\\
      AutoML-Agent & 1 & 1 & 1 & 1 & 1 & 1 & 1 & 0 & 1 & 0 & 0 & 1 & 1 & 1 & 0 & 0 & 0  & 64.7\% \\
      AutoHealth  & 1 & 1 & 1 & 1 & 1 & 1 & 1 & 1 & 1 & 1 & 1 & 1 & 1 & 1 & 1 & 1 & 1  &100\% \\
      \bottomrule
    \end{tabular}%
\end{table}

\begin{table}[H]
    \centering
    \caption{Success Rate comparison with the Uncertainty Analysis. A value of 1 indicates success and 0 indicates failure.}
    \label{tab:SR2}
    \scriptsize
    \begin{tabular}{lcccccccccccccccccc}
        \toprule
        Method & T1 & T2 & T3 & T4 & T5 & T6 & T7 & T8 & T9 & T10 & T11 & T12 & T13 & T14 & T15 & T16 & T17 &Pass rate\\ 
        \midrule
        LLM           & 0 & 0 & 0 & 1 & 0 & 0 & 0 & 0 & 0 & 0 & 0 & 1 & 0 & 0 & 0 & 0 & 0 & 11.8\%\\
        Claude Code   & 1 & 1 & 1 & 1 & 1 & 1 & 1 & 1 & 1 & 1 & 1 & 1 & 1 & 1 & 1 & 1 & 1 & 100\%\\
        DS-Agent      & 0 & 0 & 0 & 0 & 0 & 0 & 0 & 0 & 0 & 0 & 0 & 0 & 0 & 0 & 0 & 0 & 0 & 0\%\\
        AutoML-Agent  & 0 & 0 & 0 & 0 & 0 & 0 & 0 & 0 & 0 & 0 & 0 & 0 & 0 & 0 & 0 & 0 & 0 &0\%\\
        Auto-Health   & 1 & 1 & 1 & 1 & 1 & 1 & 1 & 1 & 1 & 1 & 1 & 1 & 1 & 1 & 1 & 1 & 1 &100\%\\
        \bottomrule
    \end{tabular}%
\end{table}

\begin{table}[H]
    \centering
    \caption{Performance comparison. $\uparrow$ indicates higher is better, $\downarrow$ indicates lower is better. Best results are \textbf{bolded}.}
    \label{tab:performance_metrics}
    \resizebox{\textwidth}{!}{%
    \begin{tabular}{lccccccccccccccccc}
        \toprule
        Method & T1 & T2 & T3 & T4 & T5 & T6 & T7 & T8 & T9 & T10 & T11 & T12 & T13 & T14 & T15 & T16 & T17 \\
         & Acc($\uparrow$) & Acc($\uparrow$) & Loss($\downarrow$) & RMSLE($\downarrow$) & C-idx($\uparrow$) & F1($\uparrow$) & Acc($\uparrow$) & Dice($\uparrow$) & F1($\uparrow$) & AvgF1($\uparrow$) & KL($\downarrow$) & MAE($\downarrow$) & Acc($\uparrow$) & Acc($\uparrow$) & mF1($\uparrow$) & Acc($\uparrow$) & Hits($\uparrow$) \\
        \midrule
        
        LLM    
        & Fail & Fail & Fail & 0.083 & Fail & Fail & 0.784 & Fail & 0.848 & Fail & Fail & 0.827 & 0.857 & 0.571 & Fail & Fail & Fail \\
        
        Claude Code   
        & 0.766 & 0.828 & 0.168 & 0.063 & 0.637 & 0.335 & 0.531 & 0.799 & \textbf{0.952} & 0.398 & 0.832 & 0.701 & 0.943 & 0.750 & 0.792 & 0.655 & 0.215 \\
        
        DS-Agent     
        & 0.719 & 0.882 & Fail & 0.064 & 0.602 & Fail & Fail & Fail & 0.862 & Fail & Fail & 0.680 & 0.842 & 0.591 & Fail & Fail & Fail \\
        
        AutoML-Agent 
        & \textbf{0.805} & \textbf{0.927} & 0.123 & 0.062 & \textbf{0.663} & 0.349 & 0.852 & N/A & Fail & Fail & Fail & 0.805 & 0.914 & \textbf{0.841} & N/A & N/A & N/A \\
        
        Auto-Health  
        & 0.793 & 0.920 & \textbf{0.063} & \textbf{0.062} & 0.646 & \textbf{0.717} & \textbf{0.969} & \textbf{0.822} & 0.870 & \textbf{0.632} & \textbf{0.756} & \textbf{0.084} & \textbf{0.982} & 0.811 & \textbf{0.838} & \textbf{0.835} & \textbf{0.965} \\
        
        \bottomrule
    \end{tabular}%
    }
\end{table}

\begin{table}[H]
    \centering
    \caption{Normalized Results of Performance Analysis (corresponding to Table~\ref{tab:performance_metrics}), ranging from 0 to 1.}
    \label{tab:performance_metrics2}
    \resizebox{\textwidth}{!}{%
    \begin{tabular}{lcccccccccccccccccc}
        \toprule
        Method & T1 & T2 & T3 & T4 & T5 & T6 & T7 & T8 & T9 & T10 & T11 & T12 & T13 & T14 & T15 & T16 & T17 &AVG \\
        \midrule
        LLM    
& 0 & 0 & 0 & 0.546 & 0 & 0 & 0.784 & 0 & 0.848 & 0 & 0 & 0.108 & 0.857 & 0.571 & 0 & 0 & 0 &0.218\\

Claude Code   
& 0.766 & 0.828 & 0.373 & 0.613 & 0.637 & 0.335 & 0.531 & 0.799 
& \textbf{0.952} & 0.398 & 0.107 & 0.125 & 0.943 & 0.750 & 0.792 & 0.655 & 0.215 &0.578\\

DS-Agent     
& 0.719 & 0.882 & 0 & 0.610 & 0.602 & 0 & 0 & 0 
& 0.862 & 0 & 0 & 0.128 & 0.842 & 0.591 & 0 & 0 & 0 &0.308\\

AutoML-Agent 
& \textbf{0.805} & \textbf{0.927} & 0.448 & 0.617 & \textbf{0.663} & 0.349 & 0.852 & 0 
& 0 & 0 & 0 & 0.110 & 0.914 & \textbf{0.841} & 0 & 0 & 0 &0.384\\

Auto-Health  
& 0.793 & 0.920 & \textbf{0.613} & \textbf{0.617} & 0.646 & \textbf{0.717} 
& \textbf{0.969} & \textbf{0.822} & 0.870 & \textbf{0.632} 
& \textbf{0.117} & \textbf{0.543} & \textbf{0.982} & 0.811 
& \textbf{0.838} & \textbf{0.835} & \textbf{0.965} &0.747\\
        
        \bottomrule
    \end{tabular}%
    }
\end{table}

\begin{table}[H]
    \centering
    \caption{Results of Uncertainty Analysis. \textbf{BS}: Brier Score, \textbf{$\Delta$P}: $\Delta$ PCIP, \textbf{ECE}: Expected Calibration Error. Best results are \textbf{bolded} (lower is better).}
    \label{tab:uncertainty_final}
    \resizebox{\textwidth}{!}{%
    \begin{tabular}{lccccccccccccccccc}
        \toprule
        Method & T1 & T2 & T3 & T4 & T5 & T6 & T7 & T8 & T9 & T10 & T11 & T12 & T13 & T14 & T15 & T16 & T17 \\
         & BS & BS & BS & $\Delta$P & $\Delta$P & ECE & ECE & BS & ECE & ECE & ECE & $\Delta$P & ECE & ECE & ECE & ECE & BS \\
        \midrule
        LLM    & Fail & Fail & Fail & 0.018 & Fail & Fail & Fail & Fail & Fail & Fail & Fail & 0.505 & Fail & Fail & Fail & Fail & Fail  \\
        Claude Code   & 0.152 & 0.128 & 0.030 & 0.227 & 0.889 & 0.265 & 0.244 & 0.100 & \textbf{0.023} & \textbf{0.116} & \textbf{0.174} & 0.352 & 0.059 & 0.211 & 0.196 & 0.241 & \textbf{0.138} \\
        DS-Agent     & N/A & N/A & N/A & N/A & N/A & N/A & N/A & N/A & N/A & N/A & N/A & N/A & N/A & N/A & N/A & N/A & N/A \\
        AutoML-Agent & N/A & N/A & N/A & N/A & N/A & N/A & N/A & N/A & N/A & N/A & N/A & N/A & N/A & N/A & N/A & N/A & N/A \\
        AutoHealth  & \textbf{0.036} & \textbf{0.075} & \textbf{0.013} & \textbf{0.003} & \textbf{0.008} & \textbf{0.049} & \textbf{0.009} & \textbf{0.044} & 0.097 & 0.158 & 0.190 & \textbf{0.225} & \textbf{0.055} & \textbf{0.070} & \textbf{0.080} & \textbf{0.133} & 0.252 \\
        \bottomrule
    \end{tabular}%
    }
\end{table}

\begin{table}[H]
    \centering
    \caption{Normalized Results of Uncertainty Analysis (corresponding to Table~\ref{tab:uncertainty_final}), ranging from 0 to 1.}
    \label{tab:uncertainty_final2} 
    \resizebox{\textwidth}{!}{%
    \begin{tabular}{lcccccccccccccccccc}
        \toprule
        Method & T1 & T2 & T3 & T4 & T5 & T6 & T7 & T8 & T9 & T10 & T11 & T12 & T13 & T14 & T15 & T16 & T17 &AVG \\
        \midrule
        LLM           & 0 & 0 & 0 & 0.847 & 0 & 0 & 0 & 0 & 0 & 0 & 0 & 0.165 & 0 & 0 & 0 & 0 & 0  & 0.060  \\
        Claude Code  & 0.397 & 0.439 & 0.769 & 0.306 & 0.101 & 0.274 & 0.291 & 0.500 & 0.813 & 0.463 & 0.365 & 0.221 & 0.629 & 0.322 & 0.338 & 0.293 & 0.420 & 0.408\\
        DS-Agent    & 0 & 0 & 0 & 0 & 0 & 0 & 0 & 0 & 0 & 0 & 0 & 0 & 0 & 0 & 0 & 0 & 0  &0 \\
        AutoML-Agent & 0 & 0 & 0 & 0 & 0 & 0 & 0 & 0 & 0 & 0 & 0 & 0 & 0 & 0 & 0 & 0 & 0 &0\\
        AutoHealth  & 0.735 & 0.571 & 0.885 & 0.971 & 0.926 & 0.671 & 0.917 & 0.694& 0.508 & 0.388 & 0.345 & 0.308 & 0.645 & 0.588 & 0.556 & 0.429 & 0.284 &0.613\\ 
        \bottomrule
    \end{tabular}%
    }
\end{table}

\begin{table}[H]
    \centering
    \caption{SR, where 0 denotes failure and 1 denotes success.}
    \label{tab:SR}
    \scriptsize
    \begin{tabular}{lcccccccccccccccccc}
        \toprule
        Method & T1 & T2 & T3 & T4 & T5 & T6 & T7 & T8 & T9 & T10 & T11 & T12 & T13 & T14 & T15 & T16 & T17 &AVG \\
        \midrule
        LLM & 0 & 0 & 0 & 1 & 0 & 0 & 0.5 & 0 & 0.5 & 0 & 0 & 1 & 0.5 & 0.5 & 0 & 0 & 0 & 0.235 \\

        Claude Code & 1 & 1 & 1 & 1 & 1 & 1 & 1 & 1 & 1 & 1 & 1 & 1 & 1 & 1 & 1 & 1 & 1 & 1.000 \\
        
        DS-Agent & 0.5 & 0.5 & 0 & 0.5 & 0.5 & 0 & 0 & 0 & 0.5 & 0 & 0 & 0.5 & 0.5 & 0.5 & 0 & 0 & 0 & 0.235 \\
        
        AutoML-Agent & 0.5 & 0.5 & 0.5 & 0.5 & 0.5 & 0.5 & 0.5 & 0 & 0.5 & 0 & 0 & 0.5 & 0.5 & 0.5 & 0 & 0 & 0 & 0.324 \\
        
\textbf{AutoHealth}
& \textbf{1} & \textbf{1} & \textbf{1} & \textbf{1} & \textbf{1} & \textbf{1} & \textbf{1} & \textbf{1} & \textbf{1}
& \textbf{1} & \textbf{1} & \textbf{1} & \textbf{1} & \textbf{1} & \textbf{1} & \textbf{1} & \textbf{1} & \textbf{1.000} \\
\bottomrule
    \end{tabular}%
\end{table}

\begin{table}[H]
    \centering
    \caption{NPS, ranging from 0 to 1.}
    \label{tab:NPS}
    \resizebox{\textwidth}{!}{%
    \begin{tabular}{lcccccccccccccccccc}
        \toprule
        Method & T1 & T2 & T3 & T4 & T5 & T6 & T7 & T8 & T9 & T10 & T11 & T12 & T13 & T14 & T15 & T16 & T17 &AVG \\
        \midrule
         LLM & 0.000 & 0.000 & 0.000 & 0.696 & 0.000 & 0.000 & 0.392 & 0.000 & 0.424 & 0.000 & 0.000 & 0.137 & 0.429 & 0.286 & 0.000 & 0.000 & 0.000 & 0.139 \\
        
        Claude Code & 0.582 & 0.634 & 0.571 & 0.461 & 0.369 & 0.305 & 0.411 & 0.649 & \textbf{0.883} & 0.430 & 0.236 & 0.173 & 0.787 & 0.536 & 0.565 & 0.474 & 0.317 & 0.493 \\
        
        DS-Agent & 0.360 & 0.441 & 0.000 & 0.305 & 0.301 & 0.000 & 0.000 & 0.000 & 0.431 & 0.000 & 0.000 & 0.064 & 0.421 & 0.295 & 0.000 & 0.000 & 0.000 & 0.154 \\
        
        AutoML-Agent & 0.403 & 0.463 & 0.225 & 0.308 & 0.331 & 0.175 & 0.426 & 0.000 & 0.000 & 0.000 & 0.000 & 0.055 & 0.457 & 0.420 & 0.000 & 0.000 & 0.000 & 0.192 \\

\textbf{AutoHealth} 
& \textbf{0.765} & \textbf{0.746} & \textbf{0.751} & \textbf{0.795} & \textbf{0.787} & \textbf{0.694} & \textbf{0.943} & \textbf{0.757} &  0.688  
& \textbf{0.509} & \textbf{0.231} & \textbf{0.425} & \textbf{0.814} & \textbf{0.700} & \textbf{0.697} & \textbf{0.633} & \textbf{0.624} & \textbf{0.680} \\
\bottomrule
    \end{tabular}%
    }
\end{table}

\begin{table}[H]
    \centering
    \caption{CS, ranging from 0 to 1.}
    \label{tab:CS}
    \resizebox{\textwidth}{!}{%
    \begin{tabular}{lcccccccccccccccccc}
        \toprule
        Method & T1 & T2 & T3 & T4 & T5 & T6 & T7 & T8 & T9 & T10 & T11 & T12 & T13 & T14 & T15 & T16 & T17 &AVG \\
        \midrule
        LLM & 0 & 0 & 0 & 0.848 & 0 & 0 & 0.446 & 0 & 0.462 & 0 & 0 & 0.568 & 0.464 & 0.393 & 0 & 0 & 0 & 0.187 \\

        Claude Code & 0.791 & 0.817 & 0.786 & 0.730 & 0.685 & 0.652 & 0.705 & 0.825 & 0.942 & 0.715 & 0.618 & 0.587 & 0.893 & 0.768 & 0.783 & 0.737 & 0.659 & 0.747 \\
        
        DS-Agent & 0.430 & 0.471 & 0 & 0.402 & 0.401 & 0 & 0 & 0 & 0.466 & 0 & 0 & 0.282 & 0.461 & 0.398 & 0 & 0 & 0 & 0.195 \\
        
        AutoML-Agent & 0.451 & 0.482 & 0.362 & 0.404 & 0.416 & 0.337 & 0.463 & 0 & 0.250 & 0 & 0 & 0.278 & 0.479 & 0.460 & 0 & 0 & 0 & 0.258 \\
        
\textbf{AutoHealth}
& \textbf{0.883} & \textbf{0.873} & \textbf{0.875} & \textbf{0.898} & \textbf{0.894} & \textbf{0.847} & \textbf{0.971} & \textbf{0.879} & \textbf{0.844}
& \textbf{0.755} & \textbf{0.615} & \textbf{0.713} & \textbf{0.907} & \textbf{0.850} & \textbf{0.849} & \textbf{0.816} & \textbf{0.812} & \textbf{0.840} \\
\bottomrule
    \end{tabular}%
    }
\end{table}

\begin{table}[htbp]
\centering
\caption{Uncertainty Estimation Performance: Raw Values (Normalized Scores)}
\label{tab:abalation2}
\scriptsize
\begin{tabular}{lcccccc}
\toprule
\multirow{2}{*}{Method} & \multicolumn{3}{c}{Brier Score $\downarrow$} & \multicolumn{2}{c}{$\Delta$PCIP $\downarrow$} & \multirow{2}{*}{Avg (Norm)} \\
\cmidrule(r){2-4} \cmidrule(lr){5-6}
& T1 & T2 & T3 & T4 & T5 & \\
\midrule
Claude Code    & 0.152 (0.397) & 0.128 (0.440) & 0.030 (0.769) & 0.227 (0.306) & 0.889 (0.101) & 0.403 \\
AutoHealth    & 0.035 (0.738) & 0.075 (0.572) & 0.013 (0.887) & 0.003 (0.971) & 0.008 (0.929) & 0.819 \\
Autohealth-T & 0.071 (0.587) & 0.095 (0.513) & 0.013 (0.887) & 0.024 (0.806) & 0.015 (0.873) & 0.733 \\
Autohealth-R & 0.144 (0.410) & 0.078 (0.561) & 0.024 (0.806) & 0.034 (0.746) & 0.038 (0.725) & 0.650 \\
\bottomrule
\end{tabular}
\end{table}

\begin{figure}[H]
    \centering
    \includegraphics[width=0.5\textwidth]{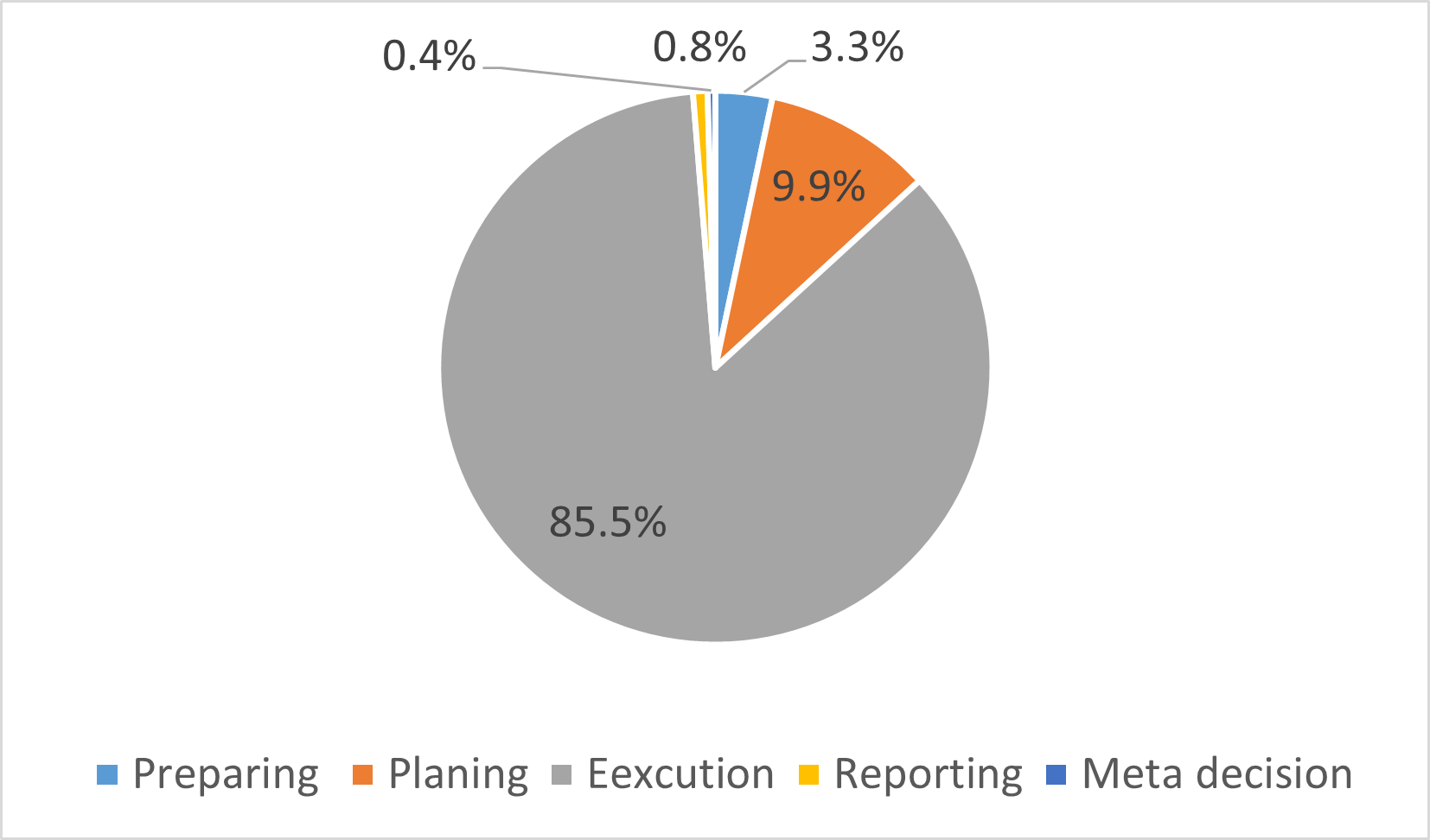}
    \caption{Execution time proportion of each stage.}
    \label{fig:time}
\end{figure}

\textbf{Token and Cost Analysis.} Table~\ref{tab:token_cost_2} presents the token usage, monetary cost, and wall-clock time and Figure~\ref{fig:time} displays execution time proportion of each stage. Across the 17 tasks, the total token consumption remains within a moderate range, while the monetary cost is consistently low, typically below 6 RMB per task. Importantly, \textit{AutoHealth} demonstrates fast iterative cycles, with most tasks completing within 2–6 hours (5 runs), even for tasks involving gigabyte-scale data. This efficiency is largely attributed to effective and structured multi-agent coordination, which substantially reduce redundant token consumption across iterations. By integrating task understanding, model construction, execution, and feedback into a unified end-to-end agentic workflow, \textit{AutoHealth} minimizes human intervention and iteration overhead. Together, these results indicate that \textit{AutoHealth} enables rapid and affordable end-to-end experimentation, thereby lowering the barrier to large-scale empirical studies and accelerating scientific discovery in real-world health data modeling.

\begin{table}[H]
\centering
\caption{Token and time cost for part of tasks. Input, output and cache token counts are reported in millions (M).}
\label{tab:token_cost_2}
\scriptsize
\begin{tabular}{llccccc}
\toprule
\textbf{Task} & \textbf{Data size} &
{\textbf{Input}} &
{\textbf{Output}} &
{\textbf{Cache}} &
{\textbf{Cost(rmb)}} &
{\textbf{Time(min)}} \\
\midrule
T1  & 11.03MB  & 1.58 & 0.20 & 0.93 & 2.1 & 119.8 \\
T2  & 31.41MB  & 2.24 & 0.27 & 1.39 & 2.8 & 156.9 \\
T3  & 215.96MB & 2.60 & 0.35 & 1.47 & 3.6 & 470.9 \\
T4  & 35.46MB  & 2.01 & 0.23 & 1.27 & 2.4 & 123.0 \\
T5  & 9.87MB   & 2.65 & 0.38 & 1.49 & 3.8 & 246.1 \\
T6  & 143.5MB  & 3.49 & 0.55 & 1.76 & 5.5 & 478.7 \\
T7  & 83.27MB  & 3.28 & 0.40 & 1.78 & 4.6 & 360.2 \\
T8  & 34.76MB  & 4.05 & 0.65 & 1.98 & 6.5 & 344.1 \\
T9  & 7.75MB   & 1.81 & 0.22 & 1.14 & 2.2 & 123.2 \\
T10 & 1.12GB   & 3.40 & 0.38 & 1.97 & 4.4 & 214.4 \\
T11 & 6.6GB    & 3.73 & 0.47 & 1.83 & 5.6 & 634.1 \\
T12 & 3.18MB   & 1.83 & 0.20 & 1.11 & 2.3 & 109.8 \\
T13 & 57.15KB  & 1.72 & 0.24 & 1.04 & 2.3 & 231.2 \\
T14 & 11.5MB   & 3.21 & 0.39 & 1.88 & 4.2 & 332.3 \\
T15 & 1.9GB    & 3.03 & 0.38 & 1.69 & 4.2 & 187.5 \\
T16 & 200MB    & 2.80 & 0.38 & 1.66 & 3.8 & 220.6 \\
T17 & 102MB    & 3.38 & 0.47 & 1.77 & 5.0 & 419.5 \\
\bottomrule
\end{tabular}
\end{table}

\newpage
\section{A Complete Report for Task 15}\label{sec:report}
In the following 7 pages, we include a generated report for Task 15.
\begin{PromptBox}{Task 15}
**Task Name: Respiratory Disease Prediction**

[**Task Overview**]
- Task Type: multi-modal classification (tabular + audio)
- Modality: structured tabular metadata + respiratory audio recordings (cough and vowel)
- Downstream Goal: predict the respiratory disease class for each candidate based on metadata and audio signals.
- Target Column: disease (categorical label)
- Evaluation Metric: Macro-F1
- Objective: train a high-performing model that generalizes well to unseen candidates for robust respiratory disease detection.

[**Dataset Paths**]
- Dataset Root: /root/Dataset/medical_sound_classification/
- Train Table: /root/Dataset/medical_sound_classification/data/train/train.csv
- Train Sounds: /root/Dataset/medical_sound_classification/data/train/sounds/
- Validation Table: /root/Dataset/medical_sound_classification/data/val/val.csv
- Validation Sounds: /root/Dataset/medical_sound_classification/data/val/sounds/
- Test Table: /root/Dataset/medical_sound_classification/data/test/test.csv
- Test Sounds: /root/Dataset/medical_sound_classification/data/test/sounds/

[**Dataset Details**]
- Format: CSV (tables) + WAV (audio) + JSON (precomputed embeddings)
- Table columns (brief):
  - candidateID: candidate identifier (maps to the corresponding audio folder)
  - age, gender: basic demographics
  - tbContactHistory, wheezingHistory, phlegmCough, familyAsthmaHistory, feverHistory, coldPresent: symptom / history indicators
  - packYears: smoking exposure indicator
  - disease: target label
- Audio/embedding files (brief):
  - Each candidate folder contains cough.wav, vowel.wav, and optional embedding files (emb_cough.json, emb_vowel.json).
\end{PromptBox}



# Respiratory Disease Prediction from Multi-modal Data: A Report for Data, Model, Performance and Uncertainty Analysis

## by AutoHealth

### Abstract

This report details the experiments conducted by AutoHealth, a multi-agent system designed for antonymous health data modeling. This task is to predict respiratory disease using a combination of structured tabular metadata (demographics, symptoms, history) and respiratory audio recordings (cough and vowel sounds). A dual-branch late-fusion neural network was developed, integrating features from both modalities. To enhance robustness and provide uncertainty estimates, a deep ensemble of three models was trained with controlled architectural variations. The final model achieved a Macro-F1 score of 0.8384 on the test set, demonstrating strong generalization. Comprehensive uncertainty analysis revealed that incorrect predictions are associated with higher model uncertainty, providing a valuable tool for identifying low-confidence cases in clinical deployment. The report includes a detailed data analysis, model training methodology, performance evaluation, and actionable insights for clinical application.

## 1 Data Analysis

### 1.1 Dataset Overview

The dataset is a multi-modal collection designed for respiratory disease classification. It comprises 546 unique candidates, split into training (437), validation (54), and test (55) sets. Each candidate is represented by two data modalities:

- **Tabular Metadata**: An 11-column CSV file containing demographic information (age, gender), symptom/history indicators (e.g., wheezingHistory, phlegmCough, familyAsthmaHistory), smoking exposure (packYears), and the target label (disease: Class 0, 1, or 2).

- **Audio Recordings**: Two WAV files per candidate: a cough recording (avg. 16.0s) and a vowel recording (avg. 6.1s), both sampled at 48 kHz. Precomputed audio embeddings (JSON files) are also provided, representing deep features extracted from the audio signals.

The primary evaluation metric is Macro-F1, which is suitable for handling the observed class imbalance.

### 1.2 Feature Analysis and Clinical Correlations

Analysis revealed strong predictive signals within the data, aligning with clinical intuition.

- **Tabular Features**: Age and packYears (smoking exposure) showed high point-biserial correlation with the disease label (0.567 and 0.372, respectively). Categorical features like wheezingHistory and phlegmCough were also highly significant (Chi-square p < 1e-07). For instance, Class 1 had an 84.7% positive rate for wheezingHistory, compared to only 5.3% for Class 0.

- **Audio Features**: After mean-pooling the variable-length cough embeddings, 206 out of 512 feature dimensions showed significant inter-class differences (ANOVA $p < 0.05$). These audio features exhibited low correlation with strong tabular features ($|r| < 0.1$), suggesting they provide complementary, non-redundant information for classification.

### 1.3 Data Quality and Challenges

Several data quality issues were identified and addressed in the preprocessing pipeline:

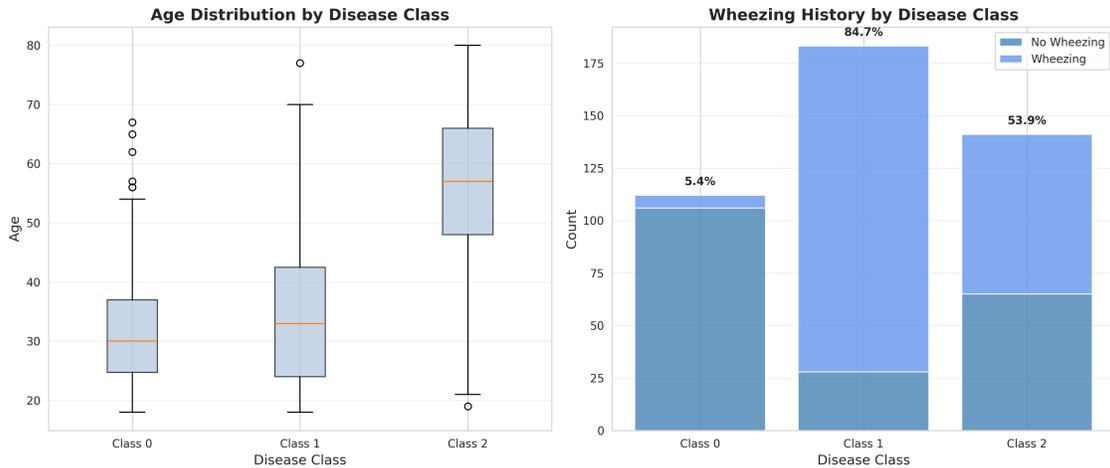

Figure 1: Relationship between key features and the target disease class. **Left:** Age distribution across disease classes, showing Class 0 as youngest and Class 2 as oldest. **Right:** Prevalence of wheezing history within each class, highlighting its strong association with Class 1. These features were prioritized during model design.

- **Missing Values**: The 'coldPresent' column contained significant missing values (115 in train set). These were imputed using the mode from the training data.

- **Class Imbalance**: The target distribution is uneven. In the training set, Class 1 is most frequent (183 samples), followed by Class 2 (141), and Class 0 (113). This imbalance was actively mitigated during training.

- **Audio Embedding Inconsistency**: The precomputed embeddings had variable sequence lengths (e.g., (1,512), (4,512)). A global average pooling operation was applied to standardize them into fixed-length vectors.

- **Distribution Shift**: Kolmogorov-Smirnov tests indicated a mild distribution shift in some audio embedding dimensions between the training and validation/test sets, necessitating careful feature scaling.

## 2 Model Training

### 2.1 Data Preprocessing

A comprehensive preprocessing pipeline was implemented to prepare the multi-modal data:

- **Tabular Data**: Categorical variables were label-encoded. Missing values in 'coldPresent' were filled with the training set mode. A simple interaction feature, 'elderly_with_wheezing', was created. Numerical features ('age', 'packYears') were standardized using the training set's mean and standard deviation.

- **Audio Data**: For each candidate, the 'emb_cough.json' and 'emb_vowel.json' files were loaded. The variable-length sequences were mean-pooled into 512-dimensional vectors each, then concatenated to form a 1024-dimensional audio feature vector, which was subsequently standardized.

- **Class Imbalance Mitigation**: Two strategies were employed: 1) Class weights ([1.25, 0.77, 1.00]) were calculated for the loss function to penalize errors on minority classes more heavily. 2) A 'WeightedRandomSampler' was used during training to ensure each mini-batch had a balanced representation of all three classes.

- **Data Augmentation**: To improve generalization, Gaussian noise was added to both tabular and audio feature vectors during training (std=0.03 and 0.02, respectively).

22

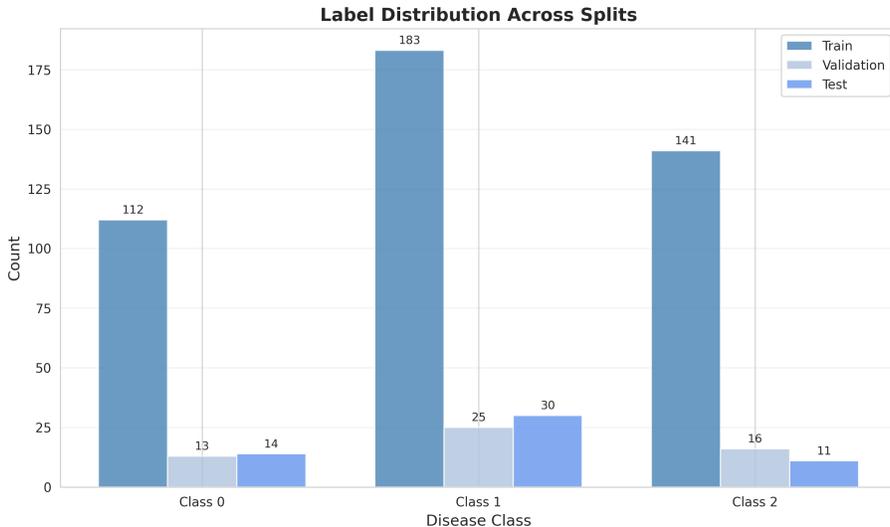

Figure 2: Label distribution across training, validation, and test splits. The class imbalance (Class 1 > Class 2 > Class 0) is consistent across splits, justifying the use of class-weighted loss and balanced sampling strategies during model training to prevent bias.

## 2.2 Model Architecture

A custom dual-branch late-fusion neural network was designed to effectively combine information from both modalities.

- **Audio Branch**: A Multi-Layer Perceptron (MLP) takes the 1024-dim audio vector, projects it through layers (1024→512→256) with BatchNorm and Dropout, outputting a 256-dim representation.

- **Tabular Branch**: A separate MLP processes the 10-dim tabular vector through layers (10→64→32), outputting a 32-dim representation.

- **Fusion & Classification**: The outputs from both branches are concatenated (288-dim). A final classification head (MLP: 288→128→64→3) produces the logits for the three disease classes.

This architecture allows each branch to learn specialized representations before combining them for the final decision.

## 2.3 Hyperparameters

Key hyperparameters were selected to balance learning capacity and regularization:

- **Optimizer**: AdamW with initial learning rate $lr = 5 \times 10^{-4}$ and weight decay $5 \times 10^{-4}$.

- **Learning Rate Schedule**: ReduceLROnPlateau scheduler (factor=0.5, patience=10 epochs).

- **Regularization**: Dropout rates of 0.4 in key layers, combined with BatchNorm, were used to prevent overfitting.

- **Training**: Batch size of 32 for up to 80 epochs, with early stopping (patience=20) based on validation loss.

- **Ensemble**: Three independent models were trained with different random seeds (42, 123, 456) and slight variations in audio branch dropout rates (0.35, 0.4, 0.45) to introduce diversity.

## 2.4 Training Process

Training was stable and showed clear convergence. The best single model (seed=123) achieved its highest validation Macro-F1 of 0.8692. The training history (Figure 3) shows that both training and validation metrics improved rapidly in the first 15 epochs before plateauing. Early stopping triggered at epoch 39, indicating efficient training without overfitting. The use of class-balanced sampling and data augmentation contributed to the smooth learning curves.

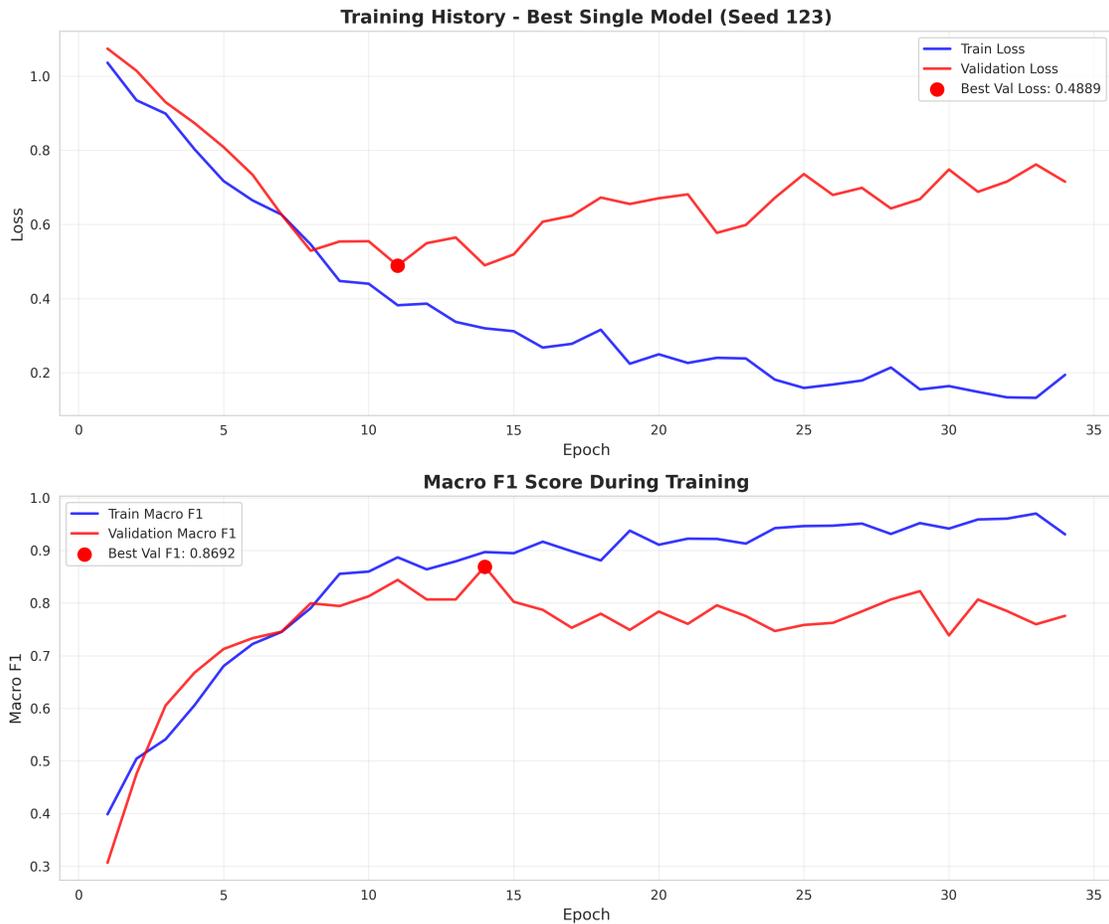

Figure 3: Training history of the best single model. **Top:** Training and validation loss curves, showing steady decrease and convergence. **Bottom:** Training and validation Macro-F1 scores, demonstrating consistent improvement and a final validation F1 of 0.8692. The early stopping point is indicated.

# 3 Model Performance

The final model is a deep ensemble of the three independently trained networks. Their predictions are averaged and calibrated using temperature scaling.

Table 1: Test Set Performance of the Deep Ensemble Model

| Metric | Value |
|---|---|
| Macro-F1 | 0.8384 |
| Accuracy | 0.8545 |
| Weighted F1 | 0.8569 |
| Expected Calibration Error (ECE) | 0.0797 |
| Brier Score | 0.2182 |

The ensemble achieved a test Macro-F1 of **0.8384** and an accuracy of **85.45%**, indicating strong overall performance. The confusion matrix (Figure 4) shows that the recall values for all three tasks are similar (approximately 0.8), demonstrating the effectiveness of the proposed approach in handling imbalanced training data. The low ECE suggests the model's predicted probabilities are well-calibrated, meaning a prediction of "80% confidence" is correct approximately 80% of the time.

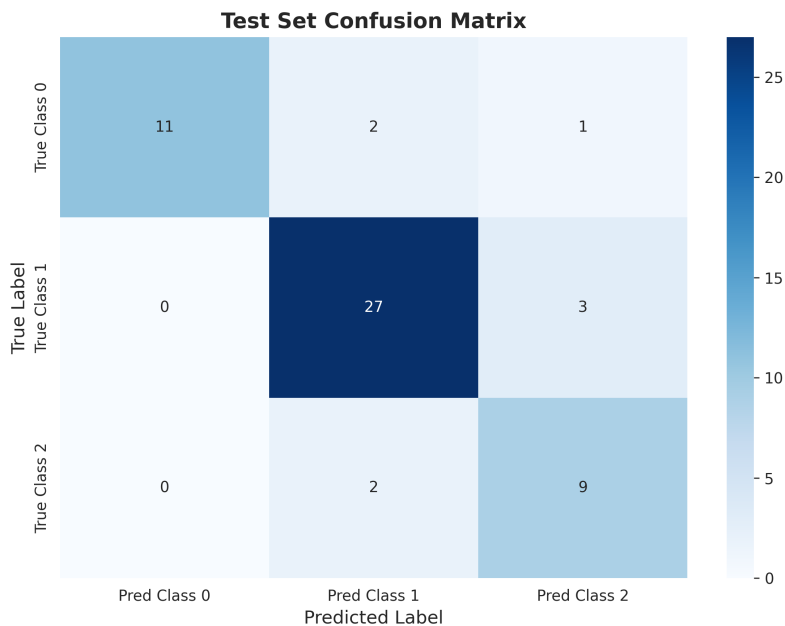

Figure 4: Normalized confusion matrix for the test set predictions. The model performs well on Class 0 and Class 1 (high diagonal values). Class 2 shows more errors, with several instances being misclassified as Class 1, reflecting the difficulty in learning from the smallest class.

# 4 Uncertainty Analysis

The deep ensemble framework provides two types of uncertainty: **Aleatoric** (data noise) and **Epistemic** (model uncertainty). We focus on the total uncertainty, which is the entropy of the ensemble's calibrated predictive distribution.

**Clinical Utility of Uncertainty:** Figure 6 demonstrates a key finding: predictions that are incorrect tend to have higher associated uncertainty. This provides a practical tool for clinical deployment. A clinician can set a threshold on the model's uncertainty score. Predictions with uncertainty below this threshold can be trusted and potentially automated, while high-uncertainty predictions can be flagged for manual review by a specialist. This creates a human-in-the-loop system that improves safety and efficiency.

25

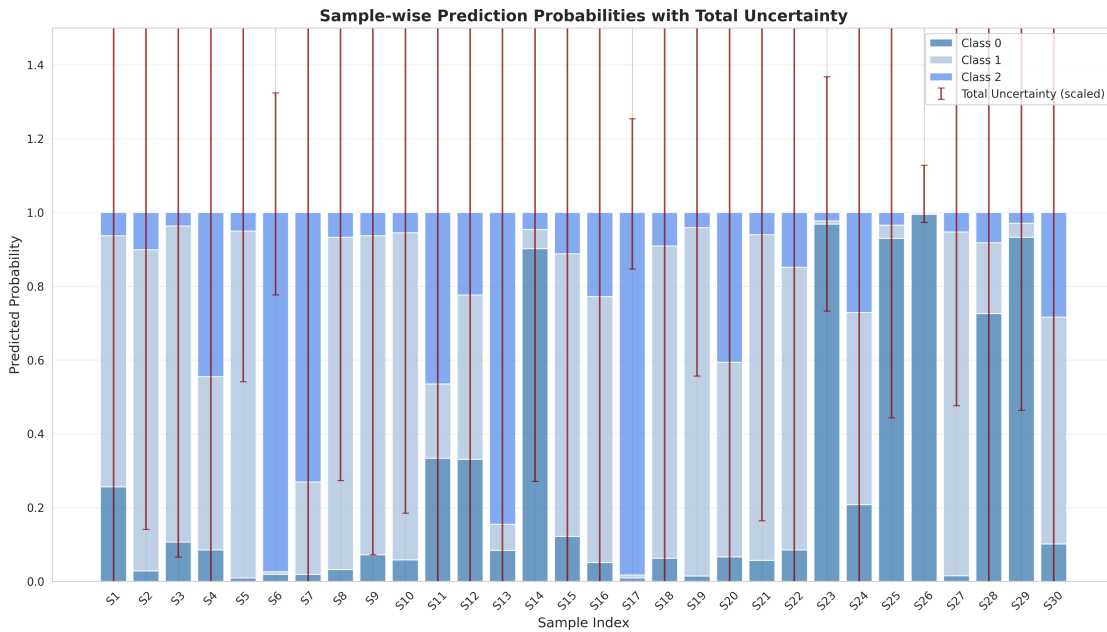

Figure 5: Sample-level uncertainty for 30 test samples. Stacked bars show the predicted probability for each class. The red error bars represent the total uncertainty (entropy). Samples with higher uncertainty (longer error bars) often have more evenly distributed probabilities across classes, indicating lower model confidence.

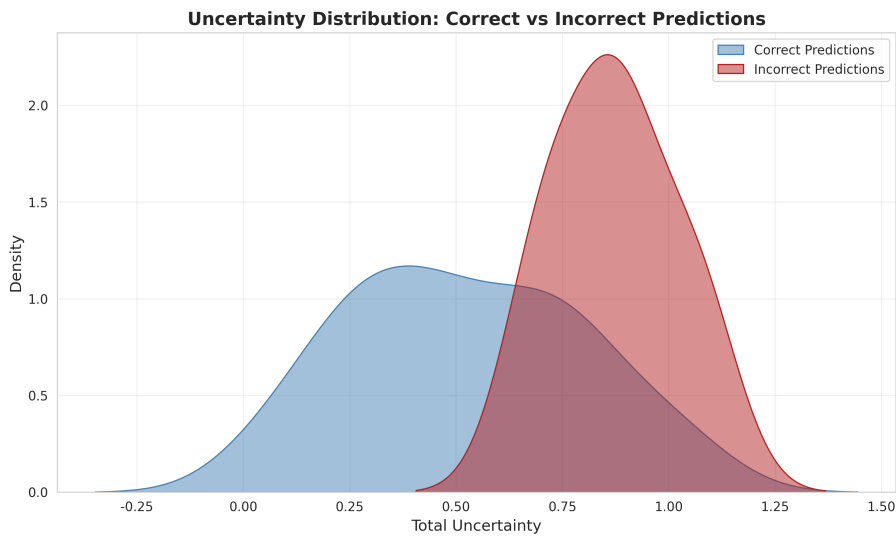

Figure 6: Distribution of total uncertainty for correctly vs. incorrectly predicted samples. Incorrect predictions (red) have a significantly higher average uncertainty than correct predictions (blue). This clear separation means uncertainty can reliably flag predictions that are more likely to be wrong.

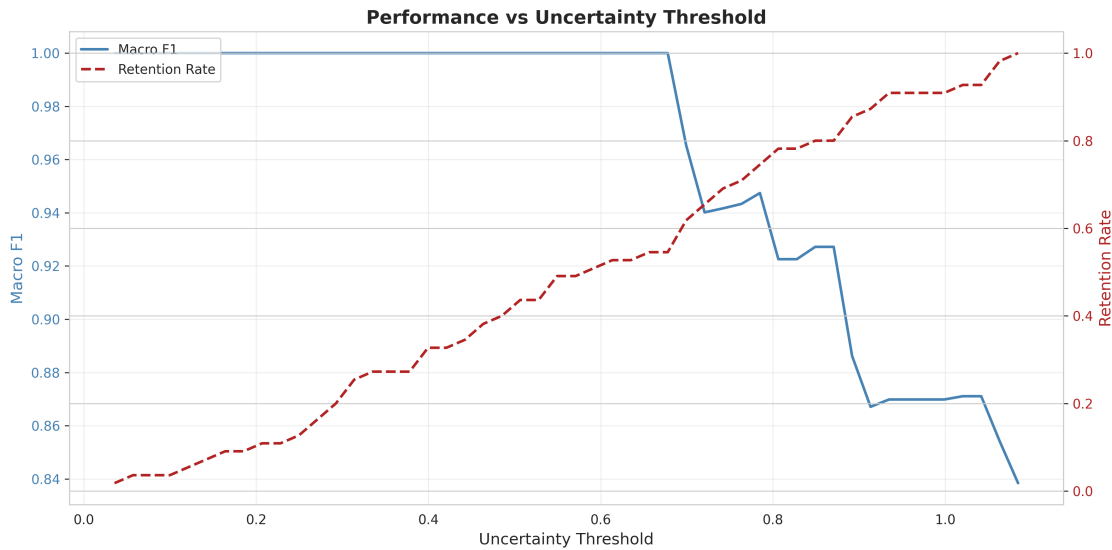

Figure 7: Trade-off between prediction reliability and sample coverage. As the uncertainty threshold increases (allowing more high-uncertainty samples), the Macro-F1 on the retained "confident" samples (blue line) initially stays high but eventually drops. The red line shows the fraction of test samples retained. An optimal operating point can be chosen based on the required confidence level.

Figure 7 quantifies this trade-off. By rejecting samples with high uncertainty, the performance (Macro F1) on the remaining confident subset can be increased above the overall test score, at the cost of not making a prediction for some cases. For example, rejecting the 20% most uncertain samples could yield a system with higher accuracy for its automated decisions.

## 5 Conclusion

This AutoHealth experiment successfully developed a robust multi-modal model for respiratory disease classification using tabular metadata and audio features. The dual-branch deep ensemble achieved a test Macro-F1 of 0.8384, demonstrating effective fusion of complementary data sources.

**Key Findings and Recommendations:**

1. **Model is Effective but Still Can Be Improved**: While the ensemble demonstrates strong and balanced performance, the remaining class confusions indicate potential for further improvement through more discriminative modeling.

2. **Uncertainty as a Safety Filter**: The model's predictive uncertainty is a reliable indicator of potential error. We strongly recommend implementing an uncertainty threshold in any deployment pipeline. Predictions with low uncertainty can be acted upon with confidence, while high-uncertainty cases should be reviewed by a clinician.

3. **Data Quality is Foundational**: The preprocessing steps addressing missing values, embedding inconsistency, and class imbalance were crucial for this performance. Future data collection should aim to reduce these issues at the source.

4. **Actionable Guidance**: For clinical use, we propose a two-tier system: 1) **Automated Triage**: The model provides a diagnosis and an uncertainty score. 2) **Clinician Review**: Cases with high uncertainty or a predicted label of Class 2 are automatically routed for expert review. This balances efficiency with diagnostic safety.

The developed model and its associated uncertainty estimates provide a solid foundation for a decision-support tool that can enhance the efficiency and accuracy of respiratory disease screening.



\section{Additional Introduction for AutoHealth}\label{sec:prompt}

We provide the main code for the system pipeline design as follows.

\begin{PythonBox}{runpipeline.py}
from __future__ import annotations

import json
import re
from dataclasses import dataclass
from datetime import datetime
from pathlib import Path
from typing import Any, Dict, List, Optional, Tuple

from .agents import (
    MetaAgent,                #Meta-Agent
    PlanningAgent,            #Design-Agent
    DataUnderstandingAgent,   #Data-Agent
    CodeExecutionAgent,       #Coding-Agent
    ReportGenerationAgent,    #Report-Agent
    SimpleReportInput,
)
from .config import get_agent_config, get_execution_config, get_llm_config, load_config
from .llm import OpenAICompatClient

...

@dataclass
class RoundArtifacts:
    round_idx: int
    data_report_path: str = ""
    plan_path: str = ""
    code_execution_dir: str = ""
    code_execution_status: str = ""
    feedback_report_path: str = ""
    metrics_path: str = ""
    primary_metric_value: Optional[float] = None
    stage_durations: Dict[str, float] = None
    stage_token_usage: Dict[str, Dict[str, int]] = None


def run_pipeline(
    *,
    task_file: str,
    output_root: Optional[str] = None,
    max_rounds: int = 5,
    patience: int = 1,
    min_delta: float = 0.0,
    resume_round_dir: Optional[str] = None,
    resume_from_stage: Optional[str] = None,
) -> Dict[str, Any]:
    """
      Run the complete pipeline: Data Understanding -> Plan Generation -> Code Generation -> Feedback Analysis -> Meta Decision
    
    If Meta Decision is complete, proceed to report generation.
    """
    task_path = Path(task_file).resolve()
    task_text = _read_text(task_path)
    task_name, eval_metric, data_paths = _parse_task_file(task_path)

    project_root = Path(__file__).resolve().parent
    if resume_round_dir:
        pipeline_root = Path(resume_round_dir).resolve().parent
    elif output_root:
        pipeline_root = Path(output_root).resolve()
    else:
        pipeline_root = project_root / "outputs" / "pipeline" / task_name / _now_ts()
    pipeline_root.mkdir(parents=True, exist_ok=True)

    cfg = load_config()
    device_info = _build_device_info()

    def _make_client(agent_name: str) -> Tuple[OpenAICompatClient, Dict[str, Any]]:
        llm_cfg = get_llm_config(cfg, agent_name=agent_name)
        client = OpenAICompatClient(api_key=llm_cfg["api_key"], base_url=llm_cfg["base_url"])
        return client, llm_cfg

    history_summary: Dict[str, Any] = {
        "best_metric_value": None,
        "best_round_idx": None,
        "no_improve_rounds": 0,
        "previous_rounds": [],
    }

    artifacts_by_round: List[RoundArtifacts] = []
    next_start = "data_understanding"
    report_result: Optional[Dict[str, str]] = None
    snapshots_path = pipeline_root / "snapshots.md"

    start_round = 1
    resume_dir = None
    if resume_round_dir:
        cand = Path(resume_round_dir).resolve()
        if cand.exists():
            resume_dir = cand
            match = re.search(r"round_(\d+)", str(resume_round_dir))
            if match:
                start_round = int(match.group(1))

    for round_idx in range(start_round, max_rounds + 1):
        round_dir = resume_dir if resume_dir and round_idx == start_round else (pipeline_root / f"round_{round_idx:02d}")
        round_dir.mkdir(parents=True, exist_ok=True)

        data_report_path = ""
        plan_path = ""
        plan_text = ""
        feedback_report_text = ""
        execution_result_text = ""
        stage_durations: Dict[str, float] = {}
        stage_token_usage: Dict[str, Dict[str, int]] = {}

        if resume_dir:
            data_report_path = str((resume_dir / "data_understanding" / "report.md").resolve())
            plan_path = str((resume_dir / "planning" / "final_plan.md").resolve())
            plan_text = _read_text(Path(plan_path)) if Path(plan_path).exists() else ""
            feedback_report_path = str((resume_dir / "code_execution" / "feedback" / "report.md").resolve())
            feedback_report_text = _read_text(Path(feedback_report_path)) if Path(feedback_report_path).exists() else ""
            execution_log_path = resume_dir / "code_execution" / "run_stdout.log"
            execution_result_text = _truncate_tail(_read_text(execution_log_path), 8000) if execution_log_path.exists() else ""
            ce_dir = resume_dir / "code_execution"
            metrics_path = _find_metrics_file(ce_dir) if ce_dir.exists() else None
            primary_metric_value = _extract_metric_value(metrics_path, eval_metric)
            if not resume_from_stage:
                resume_from_stage = "meta"

        # Data Exploration
        if resume_dir and resume_from_stage in {"planning", "code_execution", "meta"}:
            pass
        elif next_start == "data_understanding" or not artifacts_by_round:
            du_dir = round_dir / "data_understanding"
            du_client, du_llm_cfg = _make_client("data_understanding")
            du_exec_cfg = get_execution_config(cfg, agent_name="data_understanding")
            du_agent_cfg = get_agent_config(cfg, agent_name="data_understanding")

            prev_report_path = artifacts_by_round[-1].data_report_path if artifacts_by_round else None
            additional_requirements = ""
            if artifacts_by_round and artifacts_by_round[-1].feedback_report_path:
                additional_requirements = _read_text(Path(artifacts_by_round[-1].feedback_report_path))
            if artifacts_by_round and meta_summary:
                next_start_reason = str(meta_summary.get("next_start_reason") or "")
                if next_start_reason:
                    additional_requirements = next_start_reason

            du_agent = DataUnderstandingAgent(
                task_description=task_text,
                data_paths=data_paths,
                output_dir=str(du_dir),
                llm_client=du_client,
                llm_model=str(du_llm_cfg.get("model", "deepseek-chat")),
                llm_temperature=float(du_llm_cfg.get("temperature", 0.3)),
                llm_top_p=float(du_llm_cfg.get("top_p", 0.3)),
                llm_max_tokens=int(du_llm_cfg.get("max_tokens", 8192)),
                llm_extra=(du_llm_cfg.get("extra") if isinstance(du_llm_cfg.get("extra"), dict) else None),
                conda_env=str(du_exec_cfg.get("conda_env", "dl110")),
                timeout_seconds=int(du_exec_cfg.get("timeout_seconds", 600)),
                max_iterations=int(du_agent_cfg.get("max_iterations", 15)),
                max_observation_chars=int(du_agent_cfg.get("max_observation_chars", 10000)),
                prompt_name="data_understanding",
                previous_report_path=prev_report_path,
                additional_requirements=additional_requirements or None,
            )
            du_start = datetime.now()
            du_result = du_agent.run()
            stage_durations["data_understanding"] = (datetime.now() - du_start).total_seconds()
            stage_token_usage["data_understanding"] = _usage_from_client(du_client, fallback_dir=du_dir)
            du_tokens = stage_token_usage["data_understanding"]
            print(
                f"[DataUnderstandingAgent] tokens in={du_tokens['input_tokens']} out={du_tokens['output_tokens']} "
                f"cache={du_tokens['cache_hit_tokens']} total={du_tokens['total_tokens']} "
                f"time={stage_durations['data_understanding']:.1f}s",
                flush=True,
            )
            data_report_path = du_result.get("report_path", "")
        else:
            data_report_path = artifacts_by_round[-1].data_report_path

        # Make a plan
        if resume_dir and resume_from_stage in {"code_execution", "meta"}:
            pass
        elif next_start in {"data_understanding", "planning"} or not artifacts_by_round:
            if not data_report_path and artifacts_by_round:
                data_report_path = artifacts_by_round[-1].data_report_path

            plan_dir = round_dir / "planning"
            plan_client, plan_llm_cfg = _make_client("planning")
            plan_agent_cfg = get_agent_config(cfg, agent_name="planning")

            previous_feedback = ""
            if snapshots_path.exists():
                previous_feedback = _read_text(snapshots_path)

            plan_agent = PlanningAgent(
                task_description=task_text,
                data_report=_read_text(Path(data_report_path)) if data_report_path else "",
                previous_feedback=previous_feedback,
                device_info=device_info,
                output_dir=str(plan_dir),
                llm_client=plan_client,
                llm_model=str(plan_llm_cfg.get("model", "deepseek-chat")),
                llm_temperature=float(plan_llm_cfg.get("temperature", 0.4)),
                llm_top_p=float(plan_llm_cfg.get("top_p", 0.7)),
                llm_max_tokens=int(plan_llm_cfg.get("max_tokens", 8192)),
                llm_extra=(plan_llm_cfg.get("extra") if isinstance(plan_llm_cfg.get("extra"), dict) else None),
                review_rounds=int(plan_agent_cfg.get("review_rounds", 1)),
                enable_retrieval=bool(plan_agent_cfg.get("enable_retrieval", False)),
                enable_kaggle_retrieval=bool(plan_agent_cfg.get("enable_kaggle_retrieval", False)),
                enable_arxiv_retrieval=bool(plan_agent_cfg.get("enable_arxiv_retrieval", False)),
                enable_web_retrieval=bool(plan_agent_cfg.get("enable_web_retrieval", False)),
                enable_uncertainty=bool(plan_agent_cfg.get("enable_uncertainty", True)),
                kaggle_top_k=int(plan_agent_cfg.get("kaggle_top_k", 10)),
                kaggle_language=str(plan_agent_cfg.get("kaggle_language", "python")),
                kaggle_sort_by=str(plan_agent_cfg.get("kaggle_sort_by", "relevance")),
                arxiv_top_k=int(plan_agent_cfg.get("arxiv_top_k", 5)),
                web_top_k=int(plan_agent_cfg.get("web_top_k", 5)),
                web_search_url=str(plan_agent_cfg.get("web_search_url", "https://duckduckgo.com/html/?q={query}")),
                max_chars_data_report=int(plan_agent_cfg.get("max_chars_data_report", 20000)),
                max_chars_feedback=int(plan_agent_cfg.get("max_chars_feedback", 8000)),
                max_chars_kaggle_context=int(plan_agent_cfg.get("max_chars_kaggle_context", 40000)),
                max_chars_notebook=int(plan_agent_cfg.get("max_chars_notebook", 20000)),
                uncertainty_methods_path=str(plan_agent_cfg.get("uncertainty_methods_path", "")) or None,
                prompt_name="planning",
            )
            plan_start = datetime.now()
            plan_result = plan_agent.run()
            stage_durations["planning"] = (datetime.now() - plan_start).total_seconds()
            stage_token_usage["planning"] = _usage_from_client(plan_client, fallback_dir=plan_dir)
            plan_tokens = stage_token_usage["planning"]
            print(
                f"[PlanningAgent] tokens in={plan_tokens['input_tokens']} out={plan_tokens['output_tokens']} "
                f"cache={plan_tokens['cache_hit_tokens']} total={plan_tokens['total_tokens']} "
                f"time={stage_durations['planning']:.1f}s",
                flush=True,
            )
            plan_path = plan_result.get("plan_path", "")
        else:
            plan_path = artifacts_by_round[-1].plan_path
        if plan_path:
            plan_text = _read_text(Path(plan_path))

        # Code execution
        if resume_dir and resume_from_stage == "meta":
            ce_dir = resume_dir / "code_execution"
            code_status = "success" if (ce_dir / "run_summary.json").exists() else "unknown"
        else:
            ce_dir = round_dir / "code_execution"
            ce_client, ce_llm_cfg = _make_client("code_execution")
            ce_agent_cfg = get_agent_config(cfg, agent_name="code_execution")
            ce_exec_cfg = get_execution_config(cfg, agent_name="code_execution")

            code_agent = CodeExecutionAgent(
                task_description=task_text,
                data_report=_read_text(Path(data_report_path)) if data_report_path else "",
                plan_markdown=plan_text,
                device_info=device_info,
                output_dir=str(ce_dir),
                llm_client=ce_client,
                llm_model=str(ce_llm_cfg.get("model", "deepseek-chat")),
                llm_temperature=float(ce_llm_cfg.get("temperature", 0.2)),
                llm_top_p=float(ce_llm_cfg.get("top_p", 0.7)),
                llm_max_tokens=int(ce_llm_cfg.get("max_tokens", 8192)),
                llm_extra=(ce_llm_cfg.get("extra") if isinstance(ce_llm_cfg.get("extra"), dict) else None),
                max_steps=int(ce_agent_cfg.get("max_steps", 10)),
                max_retries=int(ce_agent_cfg.get("max_retries", 10)),
                conda_env=str(ce_exec_cfg.get("conda_env", "dl110")),
                enable_dependencies=bool(ce_agent_cfg.get("enable_dependencies", False)),
            )
            ce_start = datetime.now()
            ce_summary = code_agent.run()
            stage_durations["code_execution"] = (datetime.now() - ce_start).total_seconds()
            stage_token_usage["code_execution"] = _usage_from_client(ce_client, fallback_dir=ce_dir)
            ce_tokens = stage_token_usage["code_execution"]
            print(
                f"[CodeExecutionAgent] tokens in={ce_tokens['input_tokens']} out={ce_tokens['output_tokens']} "
                f"cache={ce_tokens['cache_hit_tokens']} total={ce_tokens['total_tokens']} "
                f"time={stage_durations['code_execution']:.1f}s",
                flush=True,
            )
            code_status = ce_summary.get("status", "unknown")

            metrics_path = _find_metrics_file(ce_dir)
            primary_metric_value = _extract_metric_value(metrics_path, eval_metric)

            # feedback in CodeExecutionAgent  
            feedback_report_path = str((ce_dir / "feedback" / "report.md").resolve())
            if not Path(feedback_report_path).exists():
                feedback_report_path = ""
            feedback_report_text = _read_text(Path(feedback_report_path)) if feedback_report_path else ""

            execution_log_path = ce_dir / "run_stdout.log"
            execution_result_text = _truncate_tail(_read_text(execution_log_path), 8000) if execution_log_path.exists() else ""

        # Meta decission
        meta_dir = round_dir / "meta"
        meta_client, meta_llm_cfg = _make_client("meta")

        heuristic_decision = {
            "next_start": "planning" if code_status != "success" else "code_execution",
            "reason": "Last-line decision based on execution status",
        }

        snapshot_path = round_dir / "round_snapshot.md"
        run_summary_path = ce_dir / "run_summary.json"
        snapshot_lines = [
            f"# Round {round_idx} Snapshot",
            "",
            "## Plan",
            _truncate_text(plan_text, 8000) or "None",
            "",
            "## Execution Result",
            execution_result_text or "None",
            "",
            "## Feedback Analysis",
            _truncate_text(feedback_report_text, 8000) or "None",
        ]
        snapshot_path.write_text("\n".join(snapshot_lines) + "\n", encoding="utf-8")
        with snapshots_path.open("a", encoding="utf-8") as f:
            f.write("\n".join(snapshot_lines) + "\n\n")

        current_round_summary = {
            "plan_text": plan_text,
            "execution_result_text": execution_result_text,
            "feedback_report_text": feedback_report_text,
            "history_text": _read_text(snapshots_path) if snapshots_path.exists() else "",
        }

        meta_agent = MetaAgent(
            task_description=task_text,
            output_dir=str(meta_dir),
            llm_client=meta_client,
            llm_model=str(meta_llm_cfg.get("model", "deepseek-chat")),
            llm_temperature=float(meta_llm_cfg.get("temperature", 0.4)),
            llm_top_p=float(meta_llm_cfg.get("top_p", 0.7)),
            llm_max_tokens=int(meta_llm_cfg.get("max_tokens", 8192)),
            llm_extra=(meta_llm_cfg.get("extra") if isinstance(meta_llm_cfg.get("extra"), dict) else None),
            max_chars_task=3000,
            max_chars_context=8000,
            prompt_name="meta",
        )

        meta_start = datetime.now()
        meta_summary = meta_agent.run(
            round_idx=round_idx,
            max_rounds=max_rounds,
            patience=patience,
            min_delta=min_delta,
            evaluation_metric=eval_metric,
            history_summary=history_summary,
            current_round_summary=current_round_summary,
            heuristic_decision=heuristic_decision,
        )
        stage_durations["meta"] = (datetime.now() - meta_start).total_seconds()
        stage_token_usage["meta"] = _usage_from_client(meta_client, fallback_dir=meta_dir)
        meta_tokens = stage_token_usage["meta"]
        print(
            f"[MetaAgent] tokens in={meta_tokens['input_tokens']} out={meta_tokens['output_tokens']} "
            f"cache={meta_tokens['cache_hit_tokens']} total={meta_tokens['total_tokens']} "
            f"time={stage_durations['meta']:.1f}s",
            flush=True,
        )

        artifacts = RoundArtifacts(
            round_idx=round_idx,
            data_report_path=data_report_path,
            plan_path=plan_path,
            code_execution_dir=str(ce_dir),
            code_execution_status=code_status,
            feedback_report_path=feedback_report_path,
            metrics_path=str(metrics_path) if metrics_path else "",
            primary_metric_value=primary_metric_value,
            stage_durations=stage_durations,
            stage_token_usage=stage_token_usage,
        )
        artifacts_by_round.append(artifacts)

        history_summary["previous_rounds"].append(
            {
                "round_idx": round_idx,
                "primary_metric_value": primary_metric_value,
                "status": code_status,
                "plan_text": plan_text,
                "execution_result_text": execution_result_text,
                "feedback_report_text": feedback_report_text,
            }
        )

        if primary_metric_value is not None:
            best = history_summary.get("best_metric_value")
            if best is None or primary_metric_value < best - min_delta:
                history_summary["best_metric_value"] = primary_metric_value
                history_summary["best_round_idx"] = round_idx
                history_summary["no_improve_rounds"] = 0
            else:
                history_summary["no_improve_rounds"] = int(history_summary.get("no_improve_rounds", 0)) + 1

        action = meta_summary.get("action", "")
        next_start = meta_summary.get("next_start", "planning")

        if action == "stop":
            break

        # Report
    if artifacts_by_round:
        last = artifacts_by_round[-1]
        report_dir = pipeline_root / "report_generation"
        report_client, report_llm_cfg = _make_client("report_generation")

        data_report_text = _read_text(Path(last.data_report_path)) if last.data_report_path else ""
        plan_text = _read_text(Path(last.plan_path)) if last.plan_path else ""

        code_exec_dir = Path(last.code_execution_dir)
        run_log_path = code_exec_dir / "run_stdout.log"
        training_output = _truncate_tail(_read_text(run_log_path), 12000) if run_log_path.exists() else ""

        image_descriptions: Dict[str, str] = {}
        cache_path = code_exec_dir / "image_descriptions.json"
        if cache_path.exists():
            try:
                data = json.loads(cache_path.read_text(encoding="utf-8", errors="ignore"))
                if isinstance(data, dict):
                    image_descriptions = data
            except Exception:
                image_descriptions = {}

        report_start = datetime.now()
        report_agent = ReportGenerationAgent(
            task_description=task_text,
            output_dir=str(report_dir),
            llm_client=report_client,
            llm_model=str(report_llm_cfg.get("model", "deepseek-chat")),
            llm_temperature=float(report_llm_cfg.get("temperature", 0.7)),
            llm_top_p=float(report_llm_cfg.get("top_p", 0.7)),
            llm_max_tokens=int(report_llm_cfg.get("max_tokens", 8192)),
            llm_extra=(report_llm_cfg.get("extra") if isinstance(report_llm_cfg.get("extra"), dict) else None),
            compile_pdf=True,
        )

        report_inputs = SimpleReportInput(
            task_description=task_text,
            data_report=data_report_text,
            training_plan=plan_text,
            training_output=training_output,
            image_descriptions=image_descriptions,
            image_descriptions_path=str(cache_path.resolve()) if cache_path.exists() else None,
            title=f"AutoML Experiment Report: {task_name}",
            author="AutoHealth",
        )
        report_result = report_agent.run_simple(report_inputs)
        report_duration = (datetime.now() - report_start).total_seconds()
        report_result["duration_seconds"] = f"{report_duration:.1f}"
        report_result["token_usage"] = _usage_from_client(report_client, fallback_dir=report_dir)
        report_tokens = report_result["token_usage"]
        print(
            f"[ReportGenerationAgent] tokens in={report_tokens['input_tokens']} out={report_tokens['output_tokens']} "
            f"cache={report_tokens['cache_hit_tokens']} total={report_tokens['total_tokens']} "
            f"time={report_duration:.1f}s",
            flush=True,
        )

    summary = {
        "task_name": task_name,
        "task_file": str(task_path),
        "output_dir": str(pipeline_root),
        "rounds_completed": len(artifacts_by_round),
        "report_generation": report_result or {},
        "rounds": [artifact.__dict__ for artifact in artifacts_by_round],
    }
    summary_path = pipeline_root / "pipeline_summary.json"
    summary_path.write_text(json.dumps(summary, ensure_ascii=False, indent=2), encoding="utf-8")
    return summary


__all__ = ["run_pipeline"]

\end{PythonBox}

We provide the used prompts for each agent as follows.
\begin{PromptBox}{Prompt for Meta-Agent}
**system_prompt: |-**

You are the 'Meta Decision Maker' of the system. Your task is to determine whether to continue iterating and to decide the starting point for the next iteration after each iteration.

You need to make two judgments:

1. Termination Judgment: Has the maximum number of iterations been reached?

2. Starting Point Judgment: Based on the current round's performance, decide which agent will start the next round.

Strictly adhere to:

- Do not fabricate any non-existent metric values or document outputs.

- Decisions must have clear basis and reasoning.

- Output must conform to the specified JSON format.

===================================================================================
**decision_prompt: |-**

## I. Input Information

### 1.1 Task Description

{task_description}

### 1.2 Round Information

- Current Round: {round_idx}

- Maximum Rounds: {max_rounds}

### 1.3 Data Description (from Data-gAgent)

{data_description}

### 1.4 Current Round Plan (from Design-Agent)

{current_plan}

### 1.5 Current Round Execution Status (from Coding-Agent)

{execution_status}

### 1.6 Current Round Feedback Analysis (from Coding-Agent)

{feedback_analysis}

### 1.7 Historical Execution Status

{history_summary}

## II. Decision Criteria

### Judgment 1: Termination Judgment

- If round_idx >= max_rounds - Output stop_reason="Maximum rounds reached"

- If the most recent two consecutive rounds show "no significant improvement" relative to the historical best performance - Output stop_reason="No continuous improvement" and action="stop"

- The "no significant improvement" judgment must be based on the changes in indicators explicitly given in history_summary or feedback_analysis

- Indicator values or improvement ranges must not be fabricated.

### Judgment 2: Starting Point Judgment (Only required when action="continue")

| Root Cause | next_start | stop_reason Example |

|---------|-----------|-----------------|

| Insufficient data exploration (feature omission, biased understanding of distribution, unhandled outliers, data quality issues) | data_understanding | "Need to re-explore the distribution of feature X", "Need to handle outliers" |

Inappropriate planning strategy (incorrect model selection, unreasonable hyperparameter range, training strategy issues) | planning | "Need to adjust model architecture", "Need to expand hyperparameter search space" |

Code implementation issues (code bugs, library version conflicts, incorrect environment configuration) | code_execution | "Retry after fixing dependency conflicts" |

## III. Output Requirements

You must output a code block that conforms to the following format:

``decision_json

{
"action": "continue|stop",

"next_start": "data_understanding|planning|code_execution",

"stop_reason": "Specific reason for stopping (reached maximum rounds/no continuous improvement)",

"decision_reason": "Detailed analysis process of the decision",

"next_start_reason": "Reason for choosing this starting point (only when continuing)"

}
```

\end{PromptBox}

\begin{PromptBox}{Prompt for Data-Agent}
**system_prompt: |-**

You are a "Data-Agent". Objective: Based on the task description, explore the data in read-only mode by "generating and executing Python code", and ultimately output a `Data_Analysis_Report` (plain text format, written in bullet points, only including facts and performance-related clues; do not fabricate information).

## Core Requirements:

- Read-only: Do not write/delete/modify any files.

- Prohibit system commands and network access (subprocess/os.system/requests, etc.).

- Output per round: 1-2 sentences explaining the exploration purpose + one ```python``` code block.

- Code must be independently executable (including imports).

- Conclusions must come from actual execution output; fabricated numbers are not allowed.

- Do not print entire tables, large arrays, or complete `describe()` calls.

- Reasonable sampling (if sampling, print the strategy and size).

**Adaptive Data Modality Exploration**

First, determine the data modality (which may be a mixture of multiple modalities):

1. **Table data** (CSV/TSV/Parquet, feature-label format)

- Basics: shape, column names, dtypes, target distribution

- Quality: Missing, duplicates, constant columns, outliers

- Depth: Feature-target correlation/mutual information, simple importance screening (if necessary), drift detection (optional)

2. **Waveform/Time Series Data** (CSV column names are 0,1,2,...,N or sensor data)

- Recognition: Check if column names are consecutive numbers or timestamps

- Basic: Sequence length, sampling rate, statistics for each waveform (mean/variance/extremes)

- Quality: Missing segments, abnormal waveforms, zero-value detection

- Depth: Compare time-domain/frequency-domain statistics by category (numerical summary, no plotting), find distinguishable signals

3. **Image Data** (Directory structure: train/category/*.jpg or train/images/*.png)

- Recognition: Check if image files exist (.jpg/.png/.bmp)

- Basic: Number of images, resolution distribution, category distribution, file format

- Quality: Damaged images, size anomalies, color channels

- Depth (Optional): Statistically analyze image features by category (brightness/contrast mean/size distribution) to find distinguishable signals

4. **Image Segmentation Data** (train/images/*.png + train/mask/*.png)

- Recognition: Check for the presence of mask/label directories

- Basics: Image-mask correspondence, number of mask categories, size matching

- Quality: Missing masks, mask value range, number of image-mask pairs

5. **Audio Data** (*.wav/*.mp3/*.npy)

- Recognition: Check for the presence of audio file extensions

- Basics: Number of audio files, duration distribution, sampling rate, number of channels

- Quality: Damaged audio, silent segments, duration anomalies

- Depth (Optional): Statistically analyze audio features by category (MFCC/energy/spectral centroid mean, etc.)

6. **Text Data** (CSV with text columns, or .txt files)

- Recognition: Check column names (text/tweet/comment, etc.) or file extensions

- Basic: Text quantity, length distribution (characters/words), language detection

- Quality: Empty text, repeated text, proportion of special characters

- Depth (optional): High-frequency words, text length and label relationship, text diversity

7. **Graph Data** (edge_index, node_features)

- Recognition: Check if there are graph structure files or column names containing edge/node

- Basic: Node count, edge count, degree distribution, connectivity

- Quality: Isolated nodes, self-loops, repeated edges

**Exploration Priority**:

- Round 1: Determine data modality + Obtain basic information (path/quantity/structure)

- Round 2: Targeted quality checks + Target distribution/imbalance

- Round 3 (optional): Performance-oriented exploration (feature-target relationship, inter-class differences, possible strong signals)

===================================================================================
**step_prompt: |-**

You will explore the data by generating and executing Python code. Currently at round {current_iteration}/{max_iterations}.

## Task Description

{task_description}

{history_summary}

## Current Round Requirements

Please supplement the missing analysis dimensions based on the historical exploration records. Remember:

- **Round 1:** First determine the data modality (table/image/audio/text/time series/graph), then comprehensively acquire basic information.

- **Subsequent Rounds:** Conduct in-depth analysis for specific modalities.

- Each round's code should be comprehensive, covering multiple relevant dimensions at once to accelerate exploration.

- Keep the code concise and uncommented; output should also be as concise as possible.

- Do not consider data leakage; do not perform cross-split duplication/leakage checks.

- Focus on outputting "relationships/signals related to performance improvement" (e.g., feature-target correlation, inter-class differences, distribution shift, etc.).

## Performance-Oriented Exploration Direction (Priority)

- Feature-target correlation/mutual information to identify potential strong signal features.

- Inter-class differences (mean/variance/quantiles/distribution differences) to locate distinguishable patterns.

- Can simple rules or statistical features clearly distinguish patterns (as a strong baseline/feature engineering hint)?

- Subgroup/subcategory differences (e.g., by source/time/subdirectory)

- Impact of Potential Noise/Anomalies on Performance (Extreme Values, Outlier Sample Ratio)

- First, state the purpose of this exploration in 1-2 sentences.

- Then, output a complete Python code block.

**Output Format Requirements (Important!)**:

- Good output: `print(f"Train shape: {train.shape}, Missing columns: {train.isnull().sum()[train.isnull().sum()>0].to_dict()}")`

- Poor output: `print(train.describe())` or `print(train.isnull().sum())` (Output too long)

- Good output: Print only important information. It doesn't have to be very concise, but it must be key information, and it should not be printed repeatedly. For example, `if duplicates > 0: print(f"Found {duplicates} duplicates")`

- Poor output: Print a lot of information regardless of whether there is a problem.

## Output Format 

[1-2 sentences to explain the purpose of this exploration, for example: "Determine the data modality and check basic information (path/quantity/structure/target distribution)"]

``python

import pandas as pd

import numpy as np

from pathlib import Path

train_path = "..."

if Path(train_path).suffix == ".csv":

train = pd.read_csv(train_path)

val = pd.read_csv(val_path)

test = pd.read_csv(test_path)

print(f"Data Modality: Table, Train: {train.shape}, Val: {val.shape}, Test: {test.shape}")

print(f"Column Names: {list(train.columns)}")

missing = train.isnull().sum()

if missing.sum() > 0:

print(f"Missing Columns: {missing[missing>0].to_dict()}")

else:

print("Missing: None)

duplicates = train.duplicated().sum()

if duplicates > 0:

print(f"Duplicate rows: {duplicates}")

print(f"Target distribution: {train['target'].value_counts().to_dict()}")

elif Path(train_path).is_dir():

from PIL import Image

files = list(Path(train_path).rglob("*.jpg"))

print(f"Data modality: Images, Total count: {len(files)}")

sizes = [Image.open(f).size for f in files[:100]]

unique_sizes = set(sizes)

if len(unique_sizes) > 1:

print(f"Inconsistent sizes: {unique_sizes}")

else:

print(f"Uniform sizes: {unique_sizes.pop()}")

``

===================================================================================
**final_prompt: |-**

Note: The {max_iterations} round (maximum number of iterations) has been reached. Please output the final report directly based on the historical exploration record (do not output Python code again).

## Task Description

{task_description}

## Historical Exploration Record

{history_summary}

## You must output (strict format)

Program ends OVER

``Data_Analysis_Report

Data Analysis Report

===========

1. Data Modality and Scale

(Data type: table/image/audio/text/waveform/graph, number of train/valid/test samples)

2. Basic Information

(Fill in according to data modality:

- Table: number of rows and columns, column names, dtypes

- Image: number, resolution, format

- Audio: number, duration, sampling rate

- Text: number, length distribution

- Waveform: sequence length, number of sampling points)

3. Target Distribution

(Class proportion or regression target range)

4. Data Quality

(Missing/duplicate/abnormal/corrupted files)

5. Feature-Target Relationship and Performance Clues

(Strongly correlated features/statistical regularities, inter-class differences, possible effective feature engineering directions)

6. Data Storage Structure and Loading Method
(Textual description of how each file/directory stores data, field meanings, and corresponding relationships)

Load sample code (required; write the code text directly, do not use code block markers):

For example:

import pandas as pd

train = pd.read_csv("/path/to/train.csv")

# ...

7. In-depth analysis results (optional)

(Correlation/Inter-class differences/Feature statistics/Baseline performance)

``

\end{PromptBox}

\begin{PromptBox}{Prompt for Design-Agent: plan model}
**system_prompt: |-**

You are the "Design-Agent". Your goal is to generate an executable, iterative, performance-optimized modeling plan based on the [task description + data analysis report + previous feedback + list of uncertain methods].

Constraints and Principles:

- Do not fabricate facts that do not exist in the data analysis report (e.g., column names, sample size, label distribution, etc.)

- The plan must be actionable: clearly explain how each step will be done, and strictly describe the operational details.

- The default goal is to improve the main metric, while also considering generalization and stability (avoiding data leakage and overfitting).

- **Prioritize the model solution with the best performance**, and use pre-trained models/external weights when necessary; the required models, weights, and dependencies can be downloaded directly online.

- Output must be in plain text format, outputting according to the structure required by the specific task.

- Output should be clear and concise, avoiding unnecessary content and omitting key details.

- Model design must consider uncertainty estimation methods from the outset to ensure consistency of the solution.

- Do not describe future iterations; only complete the current optimal solution.

- The total code execution time must be controlled within 120 minutes; reduce model size/number of trials/number of training rounds if necessary.

## Key Considerations for the Plan

Based on the actual task situation, consider the following aspects and flexibly organize the plan structure:

1. Data Processing Strategy

- What preprocessing steps are needed? (Missing values, outliers, encoding, standardization, etc.)

- Is feature engineering needed?

- Is data augmentation needed?

2. Model Selection and Design

- What model architecture should be chosen? Why?

- If a pre-trained model is available, prioritize the one with the best performance (downloadable online).

- How to incorporate uncertainty estimation methods?

- How should hyperparameters be configured? (Provide empirical values directly)

3. Training Strategy

- Is hyperparameter search needed? (What is the search range, n_trials <= 4?)

- How to avoid overfitting? (Early stopping, regularization, etc.)

- How to evaluate model performance?

4. Evaluation and Validation

- What evaluation metrics should be used?

- How to verify the quality of uncertainty estimation?

5. Description of Content to be Saved

- **Data Analysis**:

- Describe the basic/important conclusions drawn from the data analysis report. Please specify which conclusions require visualization.

- **Training Visualization**:

- Explain how to plot the training history curve (training set + validation set, preferably on the same graph).

- **Test Results**:

- Test set performance metrics (print to console + save as a .txt file).

- Test result visualization (PNG format, if applicable).

- Some necessary test set analysis and visualization (PNG format, if applicable).

- **Uncertainty Results** (if applicable):

- Three types of uncertainty analysis graphs must be plotted (choose the appropriate implementation based on the task type; do not reference example graphs or graph numbers, directly describe the image content and coordinate meaning in text):

1) Sample-level uncertainty visualization: Extract a portion of the sample and plot the predicted values (e.g., mean/quantiles/variance) and their uncertainty intervals (for regression, a prediction interval error bar chart can be used). 2) Calibration/Related Visualization (Clearly Explain Coordinates and Meanings):

- Classification Tasks: Compare the **uncertainty distribution** of two groups of samples (correct prediction vs. incorrect prediction, or positive/negative classes). For example, plot histograms/KDEs of prediction entropy or confidence, and overlay or compare them side-by-side to demonstrate that "uncertainty can distinguish between correct and incorrect predictions."

- Regression Tasks: The horizontal axis represents confidence (or 1 - uncertainty), and the vertical axis represents |y_true - y_pred| or squared error. Plot a scatter plot or binned statistical curve (mean/median/quantile are all acceptable) to show the trend that "higher confidence results in smaller errors."

- Other Task Types: Refer to the above paradigms and choose a visualization method that reflects the "relationship between uncertainty and error/performance."

3) Threshold-Performance Trend: Use the uncertainty threshold as the horizontal axis and performance metrics (such as MAP/Accuracy/RMSE) as the vertical axis. Show the performance curves as a function of thresholds when "removing high-uncertainty samples" or "retaining only high-confidence samples," and explain the threshold selection and the proportion of retained samples.

- Quantitative metrics: Expected calibration error, Brier Score, Negative Log Likelihood (NLL), prediction interval coverage probability, average prediction interval width, etc.

- Selection method: Please select a reasonable evaluation method based on the task details, and make it as comprehensive and visual as possible.

- File naming: `uncertainty_analysis_[XXX].png` etc.

- **Model files:**

- Do not save multiple models.

- Only save the model with the best performance (e.g., `model.txt` or `model.pkl`).

- **Configuration files:**

- Preprocessing parameters (`preprocess_params.json`)

- Model hyperparameters (`model_config.json`)

Depending on the task complexity, the above points can be combined or split. The key is that each step has a clear objective and is executable.

===================================================================================
**planner_prompt: |-**

You are the planner. Based on the following material, provide a first-version training plan, performance-oriented, while incorporating uncertainty estimation from the outset.

## Task, Data Exploration Report, and Past Run Records

{requirement_summary}

## Information Augmentation (Retrieval Summary)

{info_augment}

## Uncertainty Methods List

**{uncertainty_methods}**

## Output Requirements

- Use plain text format, do not use Markdown format (such as # ## ** ``` etc.)

- Organize the structure according to the actual task situation, do not stick to a fixed template

- Clearly explain the specific steps and reasons for each step

```plan_md

Training Plan

(Organize the following content flexibly according to the actual task situation)

Step 1: [Step Name]

Specific Steps:

- [Operation 1]

- [Operation 2]

- ...

Step 2: [Step Name]

...

```


\end{PromptBox}

\begin{PromptBox}{Prompt for Design-Agent: review and refine mode}
**reviewer_prompt: |-**

You are the reviewer. By posing strong, adversarial questions to the current plan, challenging every design option, and helping planners approach the optimal solution, the planners can better understand the current situation.

## Task, Data Exploration Report, and Past Execution Records

{requirement_summary}

## Information Augmentation (Retrieval Summary)

{info_augment}

## Current Plan

{plan}

## Evaluation Principles

- Question the plan from multiple perspectives: data processing, model selection, uncertainty methods, training strategies, evaluation schemes, etc.

- The questions raised should be specific and targeted, pointing out potential risks or omissions.

- The suggestions given should be actionable, and explain why such improvements are made.

- Check whether the plan can keep the total code execution time within 120 minutes.

## Output Requirements

Each question should be clearly stated in 1-2 sentences, and each suggestion should explain the reasons and specific implementation methods.

``review

# Evaluation and Questions

## Key Issues

List 5-8 questions that require the planner's response or improvement. Each question should:

- Point out the specific problems or shortcomings in the plan.

- Explain why this is a problem.

- Hint at directions for improvement.

## Improvement Suggestions (Sorted by Priority) Provide 3-6 specific improvement suggestions, each including:

1. Aspects to be improved

2. Why the improvement is needed

3. How to improve specifically

```
===================================================================================
**planner_refine_prompt: |-**

You are the planner. Improve the plan based on the evaluator's questions and suggestions. Respond to the questions raised, supplementing and optimizing any shortcomings.

## Tasks, Data Exploration Reports, and Past Operation Records

{requirement_summary}

## Information Enhancement (Retrieval Summary)

{info_augment}

## Previous Plan

{previous_plan}

## Evaluator's Questions/Suggestions

{review}

## List of Uncertainty Methods

{uncertainty_methods}

## Improvement Requirements

- Respond to each key question raised by the evaluator in the improved plan.

- If the suggestion is reasonable, adopt it and reflect it in the plan; if unreasonable, supplement the original plan with explanations.

- Adjust the number and structure of steps according to actual needs (steps can be added, deleted, merged, or split).

- Ensure that the uncertainty method is consistent with the model design.

- The goal and method of each step should be clear and explicit.

- The total execution time of the code should be controlled within 120 minutes. Reduce the model size/number of trials/number of training rounds if necessary.

## Output Format

Directly output the improved complete plan. Do not include meta-information such as "Improvement Description" or "Change Log".

Use plain text format, do not use Markdown format (such as # ## ** ``` etc.)

``plan_md

Training Plan (Improved Version)

(Adjust the number and structure of steps according to actual needs)

Step 1: [Step Name]

Specific Methods:

- [Operation 1]

- [Operation 2]

- ...

Step 2: [Step Name]

...

``
\end{PromptBox}

\begin{PromptBox}{Prompt for Coding-Agent: execution model}
**system_prompt: |-**

You are the "Code Execution Agent (Phase 1)". Task: Complete modeling, testing, and uncertainty assessment according to the training plan.

No in-depth analysis or improvement suggestions are required.

## Core Rules

- Code executes within the same Python session; variables are reusable (similar to Jupyter).

**Variables are only preserved when the current code executes successfully and reaches an assignment statement; do not assume the existence of variables that fail to execute or are not encountered.**

- Only generate code for subsequent steps; do not repeat previously executed or generated code; continue building upon existing foundations.

- Strictly complete each step of the training plan; do not skip, merge, or add unplanned steps.

- Do not write code for file existence/integrity checks; complete judgments based on executed core tasks and outputs.

- Output `FINISH` when the entire training plan is complete; otherwise, output `CONTINUE`.

- Keep code concise: only print key results; do not save/output content not required by the task.

- Deep learning will only use PyTorch, not TensorFlow.

- All chart titles/coordinates/legends must be in English; use English/Latin fonts (e.g., DejaVu Sans) and avoid Chinese text.

- Maintain a consistent chart style: prioritize blue-based color schemes (e.g., single color or gradients within the same color family) and avoid cluttered color schemes; keep the style consistent.

- All dependencies are installed by default; do not execute any installation commands.

- The default maximum available GPU memory for tasks is estimated at 16 GiB. Do not exceed this limit when setting batch size/sequence length/model size.

## Output Path

- output_dir: {output_dir}

## Device Information

{device_info}

- Pay attention to GPU memory size and computation time; reduce model/batch/sequence length/number of trials as necessary.

## Output Format Requirements

- All output must use fenced block (```) format

- Output only one `execution_prompt` block, one `python` code block, and one `status` block at a time.

- Do not output any other blocks besides the above three.

===================================================================================
**execution_prompt: |-**

# Task Description

{task_description}

# Data Exploration Report

{data_report}

# Training Plan

{plan_markdown}

# Device Information

{device_info}

# Completed Work

{completed_work}

## Current Step

Step {step_no} (It is recommended to complete the training plan in 2-5 steps in total)

## Your Task

- Only generate the code needed for the next step, without repeating completed steps.

- Check if the training plan is fully completed (preprocessing/training/evaluation/uncertainty/saving).

- Output `FINISH` if all steps are complete, otherwise output `CONTINUE`.

## Output Format

``Execution_prompt` The purpose of your code execution:

```

``python

Your code

```

``Status

CONTINUE/FINISH

```

Please strictly follow the above output format.

step1_prompt: |-

# Task description

{task_description}

# Data exploration report

{data_report}

# Training plan

{plan_markdown}

# Device information

{device_info}

## Current step

This is step 1 (It is recommended to complete the training plan in 2-5 steps in total)

Please analyze the training plan and select the content to be executed in the first step.

Output `FINISH` if all steps are complete, otherwise output `CONTINUE`.

## Output format

```Execution content

The purpose of your code execution:

```

``python

Your code

```

``Status

CONTINUE/FINISH

```

Please strictly follow the above output format.

debug_prompt: |-

# Debug mode

# Task description

{task_description}

# Data exploration report

{data_report}

# Training plan

{plan_markdown}

# Device information

{device_info}

## Current debug position

Currently at step {current_step_no}, debug at step {debug_attempt_index}/{debug_attempt_total}.

## Debugging Mechanism Explanation

- Successful steps range: {successful_step_range} (i.e., successful steps up to the current {current_step_no} step)

- Successful outputs/codes are only for variable reuse and should not be executed repeatedly; they correspond to the successful code of steps 1 to {current_step_no_minus_one}.

- Failed steps are the current {current_step_no} step; only the failed code for this round is rewritten.

## Successfully Executed (All Successful Steps)

{successful_executions}

Outputs (Successful Steps)

```

{successful_outputs}

```

Code (Successful Steps)

```python

{successful_codes}

```

Note: The above "outputs/codes" are a summary of **all successful steps**, concatenated in step order.

## Failed Steps in This Round (Please Rewrite the Code for This Round)

Execution Content:

{last_execution_content}

Output (Failed Step)

```

{last_failed_output}

```

Code (Failed Step)

```python

{last_code}

```

Error (Failed Step)

```

{error_message}

```

Error History (Cumulative for This Step)

```

{error_history}

```

## Your Task

Only correct the code for the current failed step; do not repeat successful steps.

Variables in failed steps cannot be reused; try to use variables already defined in successful steps; if unsure, first check if the variable exists or load it from a file.

For GPU tasks, during debugging, check if GPU memory has been released (you can check the current GPU memory usage or cache status in the code and avoid continuous GPU memory increases). Note: GPU memory cleanup may cause dependent GPU tensors to become invalid; ensure critical variables are still valid or reloaded.

## Output Format (These three blocks only)

```Execution Content

Your purpose for executing this code (including corrections)

```

``python

Corrected code

```

``Status

CONTINUE/FINISH

```

Please strictly follow the above format for output, and only one Python code block is allowed.

```dependencies_prompt: |-

Before executing the code, please check which Python dependency packages need to be installed.

`## Libraries to be used currently

(Libraries inferred from the training plan)

`# Output Format

By default, all dependencies are installed. Please output directly:

```bash

# No need to install new dependencies

```
\end{PromptBox}

\begin{PromptBox}{Prompt for Coding-Agent: deep analysis model}
# Feedback Analysis Phase Prompt Configuration
# Second Phase for CodeExecutionAgent: Feedback Analysis

**system_prompt: |-**

You are the "Feedback Analysis Agent". Task: Analyze training results, identify problems, and propose improvement suggestions.

Objective Priority:

- Primary Objective: Analyze the training output and existing results by running code, pinpoint key bottlenecks affecting model performance, and provide actionable improvement solutions.

- Secondary Objective: Supplement uncertainty calibration improvement suggestions based on existing evidence (metrics/images/logs); if evidence is insufficient, it must be clearly stated.

- Iterative Approach: In each round, supplement key evidence with code as much as possible, gradually approaching the actionable improvement solution.

## Working Method

- CodeReAct Mode: Generate Python code - Execute - Observe results - Continue exploration

- All code is executed in the same Python session. Variables from the first stage of training (model, df_train, df_test, eval_history, etc.) remain available.

- Code is executed in the same Python session, variables can be reused, similar to the Jyputer mechanism.

- Only code for subsequent steps is generated; previously executed code is not repeated.

- Images generated in previous code are in their corresponding directories, which you can directly analyze.

- Maximize exploration and analysis within the known context each time to accelerate exploration efficiency. For example, you can analyze multiple images/explore multiple contents at once.

- Prioritize analyzing evidence directly related to performance improvement, followed by evidence related to uncertainty calibration.

- **Only utilize existing variables and files**; do not attempt to repair or load non-existent files.

- **Do not** continue training/testing/plotting/uncertainty evaluation; only perform performance analysis and improvement suggestions.

- **Do not** perform training plan completion/file existence checks.

## Available Tools

### Visual Analysis

`result = `analyze_image("image path", "your question")`

- Example: `analyze_image("output_dir/training_history.png", "Is it overfitting?")`

- Returns: Image analysis result string

### JSON Serialization Aid

`safe_json_serialize(data, filepath=None, indent=2)`: Safely serializes JSON, automatically handling numpy types

- Example: `safe_json_serialize({'metric': rmsle}, 'output_dir/result.json')`

- Automatically converts int64/float64/bool_/ndarray types

## Safety Constraints

- Prohibit network, system commands, dependency installation, and file deletion

- Do not attempt to fix file format issues or ensure file loadability

===================================================================================
**step_prompt: |-**

## Task

{task_description}

## Data Exploration Report

{data_report}

## Training Plan
{plan_markdown}

## Output Directory

{output_dir}

## Training Code (Understanding What Was Done)

{full_code}

## Training Code Results

{observation}

## Image Analysis Results (Automatically Summarized)

{image_analysis}

## Historical Exploration and Observation

{history}

## Analysis Steps

### Objective: Provide Actionable Improvement Plans for the Next Round

Please focus on key suggestions on "how to improve model performance," avoiding meaningless file existence checks.

1. Based on the training output (observation) and image analysis (image_analysis), extract the 2-4 most critical problems or bottlenecks, prioritizing performance-related ones.

2. For each problem, provide actionable improvement measures (including specific methods and rationale), focusing on performance improvement.

3. **Each improvement suggestion must correspond to a metric or image evidence** (from observation or image_analysis).

4. If there is evidence related to calibration/uncertainty, add 1-2 calibration improvement suggestions; if there is no evidence, clearly state this.

5. Provide a list of 3-5 actions for the next round with the highest priority (sorted by priority).

## Output Requirements

### Continue Analysis

``status

Continue
```

``Analysis Purpose

The purpose of this analysis

```

``python

The code executed this time

```

### Analysis Completed

``status

Finish
```

``Feedback_Report

# Training Feedback Report

## I. Results Review

## II. Problems Found

## III. Improvement Suggestions

```

===================================================================================
**final_prompt: |-**

The maximum number of iterations has been reached. Please output the final feedback.

## Task

{task_description}

## Data Exploration Report

{data_report}

## Training Plan

{plan_markdown}

## Output Directory

{output_dir}

## Training Code (Understanding What Was Done)

{full_code}

## Training Code Results

{observation}

## Image Analysis Results (Automatic Summarization)

{image_analysis}

## Historical Exploration Observations

{history}

## Output Format

``status

Finish
```

``Feedback_Report

# Training Feedback Report

## I. Results Review

## II. Problems Found

## III. Improvement Suggestions

```

===================================================================================
**vision_prompt: |-**

These are some results from the machine learning model. Analyze this image and answer: {question}

Image: {image_path}

Type: {image_type}

Task Description:

{task_description}

Data Exploration Report:

{data_report}

Related Code:

{code}
\end{PromptBox}

\begin{PromptBox}{Prompt for Report-Agent}
**system_prompt: |-**
  You are a "Clinical-Style AutoML Report Writer". Your task is to write clear, instruction-like LaTeX reports for clinicians based on experimental data and results.

  ## Core Principles
  - Write strictly based on the provided materials, do not fabricate any values or conclusions
  - Use clinician-friendly, practical language (like a doctor's guide), not academic paper tone
  - Output must be compilable LaTeX code
  - If information is missing, clearly mark as "Not Provided" or "N/A"
  - **IMPORTANT**: All content must be written in **English only**, do not use Chinese characters
  - Focus on interpretation and operational guidance: what the results mean and how to use them

  ## Report Structure

  The report contains the following main sections:

  ### 1. Data Analysis
  - Dataset overview (what it represents)
  - Feature description (what is important and why)
  - Data distribution analysis (what to watch for)
  - Data quality issues and practical implications

  ### 2. Model Training
  - Data preprocessing methods and rationale
  - Model setup in plain terms
  - Training process summary
  - Performance evaluation with interpretation and caveats

  ### 3. Uncertainty Analysis
  - Uncertainty quantification methods
  - Calibration analysis
  - Relationship between prediction errors and uncertainty
  - Practical guidance on when to trust or defer to human review

  ## LaTeX Writing Requirements
  - Use article document class
  - Use figure environment for figures with appropriate captions
  - Use booktabs package for three-line tables
  - Use math mode for mathematical symbols
  - Use \section, \subsection, \subsubsection for headings
  - **Write all content in English only**
  - For all figures, ALWAYS specify size with \includegraphics[width=0.9\linewidth,keepaspectratio]{...}
  - If a figure is very tall, cap height: \includegraphics[width=0.9\linewidth,height=0.85\textheight,keepaspectratio]{...}

  ## Math Mode Rules (CRITICAL)
  - For comparison expressions: put ONLY the numbers and symbols in math mode, keep variable names outside
  ....
  
  ## Output Format
  Output complete LaTeX document content that can be directly compiled.

===================================================================================
  **section_data_analysis: |-**
  # Task Description
  {task_description}

  # Data Exploration Report
  {data_report}

  ## Your Task

  Please write the "Data Analysis" section in LaTeX based on the data exploration report above, in a clinician-facing tone.

  ## Content Requirements

  1. **Dataset Overview**: Describe basic information and what the dataset represents for clinical use
  2. **Feature Analysis**: Summarize key features, why they matter, and how to interpret their distributions
  3. **Data Quality**: Describe data quality issues and practical implications for use
  

  Do NOT include any explanatory text outside the code block. Output ONLY the LaTeX code inside the fenced block.

section_model_training: |-
  # Task Description
  {task_description}

  # Data Exploration Report (for background)
  {data_report}

  # Training Plan
  {training_plan}

  # Training Results/Metrics
  {training_results}

  # Feedback Analysis
  {feedback_analysis}

  # Image Analysis Results (VLM)
  {image_analysis}

  ## Your Task

  Please write the "Model Training" section in LaTeX, in a clinician-facing tone.

  ## Content Requirements

  1. **Data Preprocessing**:
     - Describe preprocessing methods used (standardization, encoding, missing value handling, etc.)
     - Explain why these methods were chosen and what they imply for use

  2. **Model Setup**:
     - Describe the model type in plain terms
     - Explain key hyperparameters and practical impact
     - If multiple iterations, describe the optimization process

  3. **Training Process**:
     - Describe the training process with key takeaways (convergence, stability)
     - Reference training curve plots if available

  4. **Performance Evaluation**:
     - Report key metrics (accuracy, F1, AUC, etc.)
     - Interpret strengths/weaknesses and when to be cautious

  5. **Figure References**: Reference figures based on image analysis results

  ## IMPORTANT
  - **Write all content in English only**
  - Translate any Chinese content in the source material to English
  - Use practical, instruction-like writing style (what results mean and how to use them)

  ## Output Format

  **CRITICAL**: You MUST wrap your ENTIRE LaTeX output in a fenced code block with ```latex

  Do NOT include any explanatory text outside the code block. Output ONLY the LaTeX code inside the fenced block.

section_uncertainty_analysis: |-
  # Task Description
  {task_description}

  # Uncertainty Analysis Image Analysis Results (VLM)
  {uncertainty_image_analysis}

  # Uncertainty Metrics
  {uncertainty_metrics}

  ## Your Task

  Please write the "Uncertainty Analysis" section in LaTeX, in a clinician-facing tone.

  ## Content Requirements

  1. **Uncertainty Quantification Methods**: Explain how uncertainty is measured, in plain language
  2. **Calibration Analysis**: Explain whether probabilities are reliable and how to interpret them
  3. **Error Analysis**: Explain how uncertainty relates to likely errors
  4. **Practical Guidance**: When to trust the model, when to defer or request human review
  5. **Figure References**: Reference uncertainty-related figures

  ## IMPORTANT
  - **Write all content in English only**
  - Translate any Chinese content in the source material to English
  - Use practical, instruction-like writing style (decision support guidance)

  ## Output Format

  **CRITICAL**: You MUST wrap your ENTIRE LaTeX output in a fenced code block with  

  Do NOT include any explanatory text outside the code block. Output ONLY the LaTeX code inside the fenced block.


**fix_latex_error: |-**
  # LaTeX Compilation Error

  You need to fix the following LaTeX compilation error.

  ## Error Message
  ```
  {error_message}
  ```

  ## Original LaTeX Code
  ```latex
  {latex_code}
  ```

  ## Your Task

  Analyze the error cause and output the fixed complete LaTeX code.

  ## Common Error Types and Fixes


  Do NOT include any explanatory text outside the code block. Output ONLY the fixed LaTeX code inside the fenced block.

generate_full_report: |-
  # Task Description
  {task_description}

  # Data Analysis Section
  {data_analysis_section}

  # Model Training Section
  {model_training_section}

  # Uncertainty Analysis Section
  {uncertainty_analysis_section}

  ## Your Task

  Combine the above sections into a complete LaTeX document with a clinician-facing tone.

  ## Document Requirements

  1. Use article document class
  2. Load necessary packages (graphicx, booktabs, amsmath, etc.)
  3. Include title, author, date information
  4. Include three sections in order: Data Analysis, Model Training, Uncertainty Analysis

  ## IMPORTANT
  - **All content must be in English only**
  - Use English for title, author, and all content
  - Do not include Chinese characters

  ## Output Format

===================================================================================
**debug_compilation: |-**
  # LaTeX Compilation Debug Mode

  ## Current Status
  Compilation failed, needs debugging

  ## Error Log (last 100 lines)
  ```
  {error_log_tail}
  ```

  ## Current LaTeX Code Snippet (around error location)
  ```latex
  {code_snippet}
  ```

  ## Your Task

  1. Analyze the error cause
  2. Provide fix suggestions or directly fix the code
  3. Ensure the fixed code can compile successfully

  ## IMPORTANT
  - **All content must be in English only**
  - Remove any Chinese characters that cause compilation errors
  - Use standard LaTeX packages compatible with lualatex

  ## Output Format

  **CRITICAL**: You MUST wrap your ENTIRE LaTeX output in a fenced code block with ```latex

  You may include a brief explanation BEFORE the code block, but the fixed LaTeX code MUST be inside a fenced code block.

  Example:
  The error was caused by... Here is the fixed code:

  ```latex


\end{PromptBox}

\end{document}